\documentclass[12pt]{article}

\usepackage{arxiv}

\usepackage[utf8]{inputenc} 
\usepackage[T1]{fontenc}    
\usepackage{hyperref}       
\usepackage{url}            
\usepackage{amsfonts}       
\usepackage{nicefrac}       
\usepackage{microtype}      
\usepackage{lipsum}
\usepackage{fancyhdr}       
\usepackage{graphicx}       
\usepackage{pifont}
\usepackage{booktabs}
\usepackage{multirow}
\usepackage[table,xcdraw]{xcolor}
\graphicspath{{Figures/}}     
\usepackage{amsmath}
\usepackage{amssymb}
\usepackage{pifont}
\title{The Computational Limits of Deep Learning
}

\author
{Neil C. Thompson$^{1\ast}$, Kristjan Greenewald$^{2}$, Keeheon Lee$^{3}$, Gabriel F. Manso$^{4}$\\
\\
\normalsize{$^{1}$MIT Computer Science and A.I. Lab,}\\ \normalsize{MIT Initiative on the Digital Economy, Cambridge, MA USA}\\
\normalsize{$^{2}$MIT-IBM Watson AI Lab, Cambridge MA, USA}\\
\normalsize{$^{3}$Underwood International College, Yonsei University, Seoul, Korea} \\
\normalsize{$^{4}$ FGA,  University of Brasilia, Brasilia, Brazil} \\
\\
\normalsize{$^\ast$To whom correspondence should be addressed; E-mail:  neil\_t@mit.edu.}
}

\begin{document}
\maketitle

\begin{abstract}
Deep learning's recent history has been one of achievement: from triumphing over humans in the game of Go to world-leading performance in image classification, voice recognition, translation, and other tasks. But this progress has come with a voracious appetite for computing power.  This article catalogs the extent of this dependency, showing that progress across a wide variety of applications is strongly reliant on increases in computing power. Extrapolating forward this reliance reveals that progress along current lines is rapidly becoming economically, technically, and environmentally unsustainable.  Thus, continued progress in these applications will require dramatically more computationally-efficient methods, which will either have to come from changes to deep learning or from moving to other machine learning methods.
\end{abstract}


\section{Introduction} \label{sec:introduction}

In this article, we present a comprehensive meta-analysis of how deep learning progress depends on growing computational power and use this to understand not just how particular models scale, but how the field as a whole does. Our analysis differs from previous ones in that we are (i) more precise in the models we compare than are many high-level historical analyses, which allows us to better understand how performance changes as computing scales up, and (ii) better able to account for innovation in the field than estimates where researchers have tested scaling by varying the compute used in training their own models.

To understand scaling in deep learning, we analyze $1{,}527$ research papers found in the arXiv pre-print repository, as well as other sources, in the domains of image classification, object detection, question answering, named entity recognition, machine translation, speech recognition, face detection, image generation, and pose estimation.  We find that computational requirements have escalated dramatically and that increases in computing power have been central to performance improvements.  

This finding has important public policy implications: if current trends continue, the growing ``computational burden'' of deep learning will rapidly become technically and economically prohibitive. Such a rapid escalation in computing needed also implies alarming growth in deep learning's environmental cost. Faced with these challenges, the machine learning community will be pushed to either dramatically increase the efficiency of deep learning\footnote{There are already significant efforts underway to increase efficiency\cite{patterson2021carbon}, as we will discuss in section \ref{sec:lessening_burden}} or to move to more computationally-efficient machine learning techniques.

To understand why deep learning is so computationally expensive, we analyze its statistical and computational scaling in theory. We show that deep learning is not computationally expensive by accident, but by design.  The same flexibility that makes it excellent at modeling diverse phenomena and outperforming expert models also makes it dramatically more computationally expensive.  Despite this, we find that the actual computational burden of deep learning models is scaling more rapidly than (known) lower bounds from theory, suggesting that substantial improvements might be possible.

It would not be a historical anomaly for deep learning to become computationally constrained.  Even at the creation of the first neural networks by Frank Rosenblatt, performance was limited by the available computation.  In the past decade, these computational constraints have been relaxed due to speed-ups from moving to specialized hardware and a willingness to invest more resources to improve performance.  But, as we show, the computational needs of deep learning scale so rapidly that they will quickly become constraining again.

\section{Deep Learning's Computational Requirements in Theory}\label{sec:theory}

In deep learning, the relationship between performance, model complexity, and computational requirements is still not well understood theoretically. 
Nevertheless, there are important reasons to believe that deep learning is intrinsically highly reliant on computing power.  This arises from the role of overparameterization and how this scales as additional training data are used to improve performance.

It has been proven that there are significant benefits to having a neural network containing more model parameters than data points available for training, that is, by \textbf{overparameterizing} it \cite{soltanolkotabi2019theoretical}. Classically this would lead to overfitting, but stochastic gradient-based optimization methods provide a \textbf{regularizing} effect due to early stopping \cite{pillaud2018statistical, Belkin15849}\footnote{This is often called \emph{implicit} regularization, since there is no explicit regularization term in the model.}, moving the neural  networks into an \emph{interpolation} regime, where the training data is fit almost exactly while still maintaining reasonable predictions on intermediate points \cite{belkin2018overfitting, belkin2019does}. An example of large-scale overparameterization is the current state-of-the-art image classification system, CoCa, which has $2.1$B parameters for imagenet's $1.2$M data points \cite{yu2022coca}.

The challenge of overparameterization is that the number of deep learning parameters must grow as the number of data points grows.  Since the cost of training a deep learning model scales with the \emph{product} of the number of parameters with the number of data points, this implies that computational requirements grow as at least the \emph{square} of the number of data points in the overparameterized setting. This quadratic scaling, however, is an underestimate of how fast deep learning networks must grow to improve \textit{performance}, because a linear improvement in performance generally requires a faster-than-linear increase in the amount of training data. 

For instance, statistical learning theory tells us that, in general, root mean squared prediction error can at most drop as $1/\sqrt{n}$ (where $n$ is the number of data points) \cite{loh2017lower}.  These rates indicate that at least a quadratic increase in data points would be needed to improve performance. So, combining the computational overhead from overparameterization and the data requirements for statistical learning yields a back-of-the envelope estimate that the computation required to train an overparameterized model should grow at least as a fourth-order polynomial with respect to performance,\footnote{Here, performance is $1/(RMSE)$.} i.e. \(Computation = \Omega(Performance^4)\).  This is, of course, just a lower bound.  Due to the complexity of deep learning, performance could be considerably worse, perhaps even requiring exponential increases in computing power as has been seen in other tasks like weather prediction \cite{thompson2020exponential}.

While the bound above was derived for root mean squared error, the result is more general, applying to the large class of performance metrics that converge as $1/\sqrt{n}$.  For example, this includes any smooth loss function (or error metric) that is computed by averaging over data points, as \cite{golowich2017size} showed.\footnote{See \cite{golowich2017size} for precise assumptions.} In particular, this result applies to most popular neural network training losses, including the cross entropy loss.

The relationship between model parameters, data, and computational requirements in deep learning can be illustrated by analogy in the setting of linear regression, where the statistical learning theory is better developed (and, which is equivalent to a 1-layer neural network with linear activations).  Under the usual conditions,\footnote{See supplement section \ref{supp:reg} for details and derivation} the root mean squared prediction error from the ordinary least-squares (OLS) estimator scales as $O\left(\sqrt{\frac{d}{n}}\right)$, where $d$ is the number of model parameters and $n$ the number of observations.  Under these conditions, and assuming stochastic gradient descent is used for estimation, learning a model with $1{,}000\times$ as many parameters would take $1{,}000{,}000\times$ longer (due to the necessary increase in $n$ to preserve the same RMSE). Regularization (either explicit regularization or the implicit regularization created by state of the art training of neural networks) can help.  For instance, the lasso estimator \cite{tibshirani1996regression}, which performs an explicit regularization, improves root mean squared error scaling to $O\left(\sqrt{\frac{s \log d}{n}}\right)$ where $s$ is the number of nonzero coefficients in the true model \cite{meinshausen2009lasso}. We make an analogy between the role of regularized lasso estimation in linear regression to the role of deep learning in nonlinear problems, since neural networks have been shown to be \emph{implicitly} regularized \cite{pillaud2018statistical,Belkin15849}.

\begin{figure}[!htb]
 \centering
 \includegraphics[width=1\textwidth]{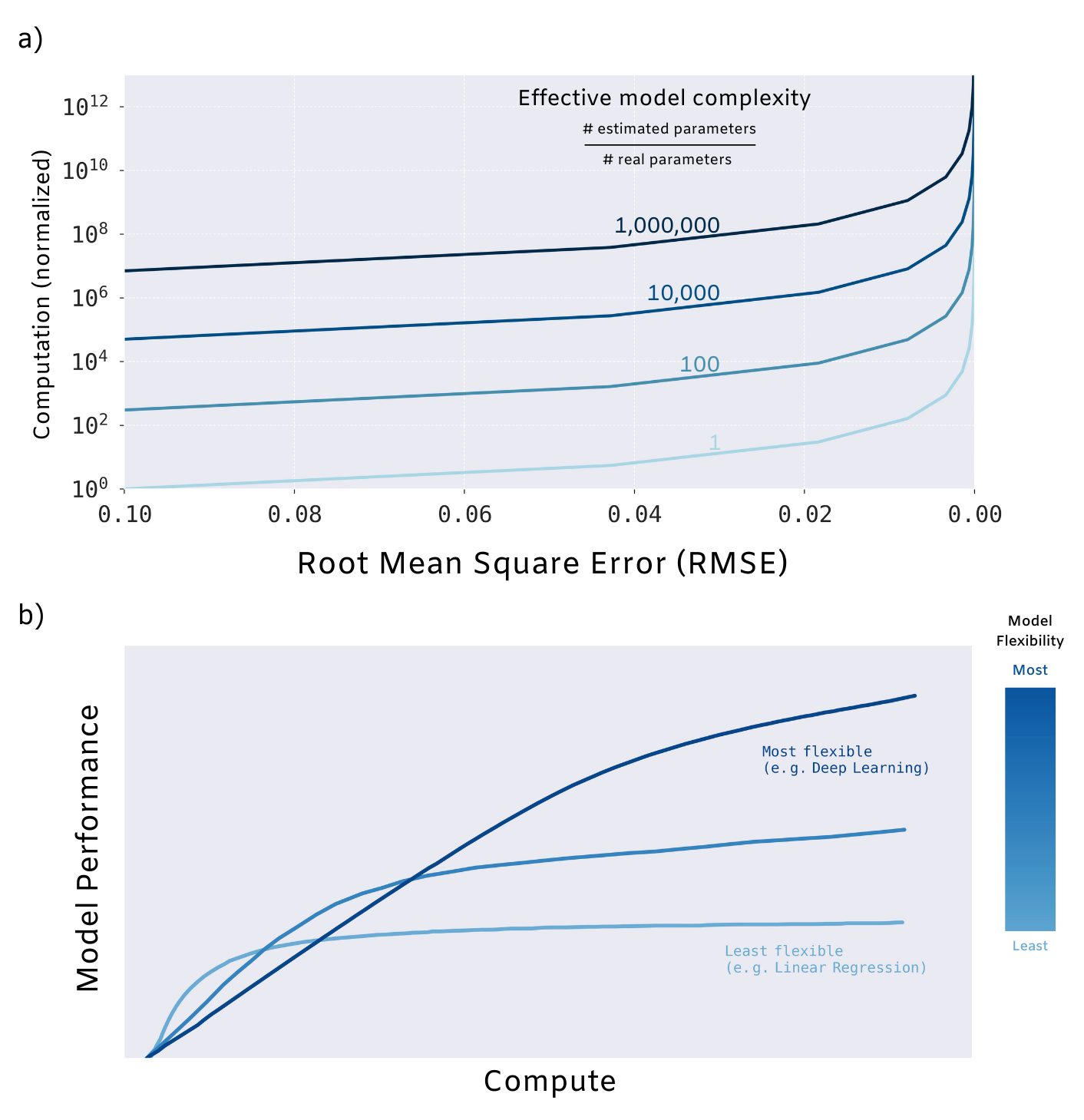}
 \caption{a) Computational burden of running regularized models (lasso) as the effective model complexity \cite{Nakkiran2019}, $d$/$s$, is increased (where $d$ is the number of parameters in the estimated model and $s$ is the number in the true model. b) Implications of flexibility for machine learning model performance.}
 \label{fig:scaling_and_ngs_graphs}
\end{figure}

Even with regularization, however, theory tells us that the computing power needed for improved performance still grows incredibly rapidly.  For example, the computational power needed to run a highly flexible (flexibility is sometimes also called ``effective model complexity'' \cite{Nakkiran2019}) lasso model with $d = 1{,}000 s$ parameters, is about ${1{,}000}\times$ that for running a lasso model with just the true number of parameters, $d=s$. 
Figure \ref{fig:scaling_and_ngs_graphs} (a) generalizes this, showing the increase in computation needed as the effective model complexity ($d/s$) increases \cite{Nakkiran2019}.

As Figure \ref{fig:scaling_and_ngs_graphs} (a) shows, there is an enormous computational price that has to be paid for building models with many parameters, even when regularization is used.  But while the price of including so many parameters may be high, it also offers flexibility for the model.  In contrast, smaller models may be more efficient, but if they do not include the parameters that matter for the answer (in the example above, some of the $s$ coefficients), this would imply lower RMSE values being unachievable for any amount of computation.  In other words, the performance of that model will eventually plateau at a low level as available computation/number of samples increase, since it lacks important predictive features. In contrast, the model with many parameters will eventually achieve a high level of performance, but at the cost of more data and computation. 

Thus, we arrive at the central tradeoff between traditional machine learning methods (like regression) that use small numbers of parameters and deep learning methods that use enormous numbers of parameters.  The more parameters that one adds to a model the greater the flexibility and hence potential for better performance.  Indeed, it has been shown that sufficiently large neural networks are \emph{universal function approximators} \cite{hornik1989multilayer}, hence in theory, any desired performance level can be achieved by making the model large enough and including enough training data.  But these additional parameters also make the model more expensive to train (even before any needed increase in amount of training data) and can make it do less well when the amount of data (or computation) is not large enough.  Figure \ref{fig:scaling_and_ngs_graphs} (b), our adaptation of a graph attributed to Andrew Ng \cite{kruup_2018}, summarizes this.

\section{Deep Learning's Computational Requirements in Practice}

\subsection{Past}\label{sec:past}

Even in their early days, it was clear that computational requirements limited what neural networks could achieve.  In $1960$, when Frank Rosenblatt wrote about a $3$-layer neural network, there were hopes that it had ``gone a long way toward demonstrating the feasibility of a perceptron as a pattern-recognizing device.''  But, as Rosenblatt already recognized ``as the number of connections in the network increases, however, the burden on a conventional digital computer soon becomes excessive'' \cite{Rosenblatt1960}.  Later that decade, in $1969$, Minsky and Papert explained the limits of $3$-layer networks, including the inability to learn the simple XOR function.  At the same time, however, they noted a potential solution: ``the experimenters discovered an interesting way to get around this difficulty by introducing longer chains of intermediate units'' (that is, by building \textit{deeper} neural networks) \cite{minsky69perceptrons}.  Despite this potential workaround, much of the academic work in this area was abandoned because there simply wasn't enough computing power available at the time. As L\'eon Bottou later wrote ``the most obvious application of the perceptron, computer vision, demands computing capabilities that far exceed what could be achieved with the technology of the $1960$s'' \cite{minsky69perceptrons}.

\begin{figure*}[!htb]
 \centering
 \includegraphics[width=1\textwidth]{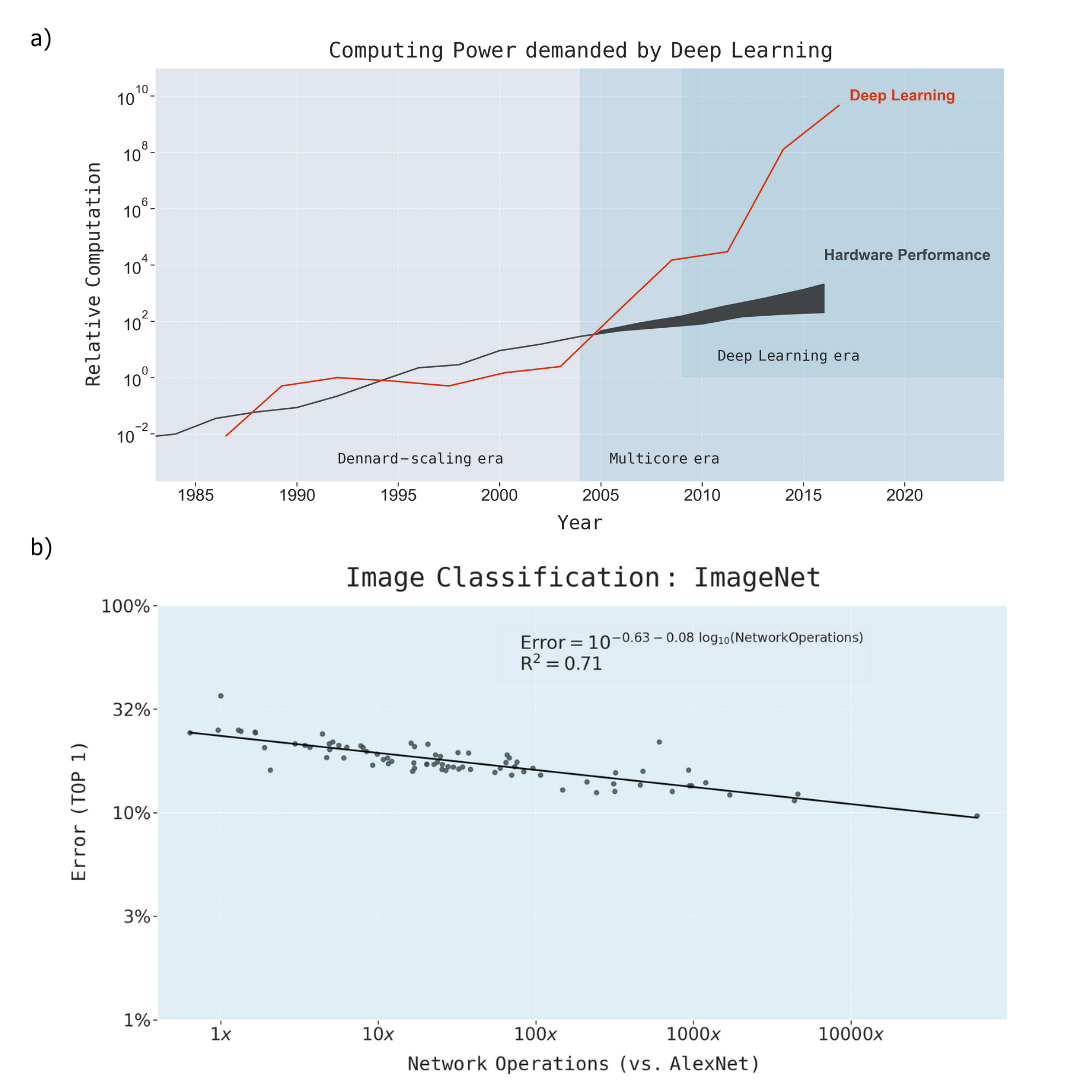}
 \caption{Computing power used in: \textbf{(a)} the largest deep learning models in different year (across all applications) \cite{openai2018} as compared with the growth in hardware performance from improving processors\cite{DanowitzKeMa12}, as analyzed by \cite{HennessyPa19a} and \cite{Leiserson20},\protect\footnotemark \textbf{(b)} image classification models tested on the ImageNet benchmark with computation normalized to the $2012$ AlexNet model \cite{krizhevsky2012imagenet}.}
 \label{fig:long_term_growth}
\end{figure*}

In the decades that followed, improvements in computer hardware provided, by one measure, an approximately $50,\!000\times$ improvement in performance \cite{HennessyPa19a} and the largest neural networks being used grew their computational requirements proportionally, as shown in Figure \ref{fig:long_term_growth}(a).  Since the growth in computing power per dollar closely mimicked the growth in computing power per chip \cite{thompson2021decline}, this meant that the economic cost of running such models was largely stable over time.  Despite this large increase,  deep learning models in $2009$ remained ``too slow for large-scale applications, forcing researchers to focus on smaller-scale models or to use fewer training examples.''\cite{Raina2009} The turning point seems to have been when deep learning was ported to GPUs, initially yielding a $5-15\times$ speed-up \cite{Raina2009} which by 2012 had grown to more than $35\times$ \cite{nvidia2017}, and which led to the important victory of Alexnet at the $2012$ Imagenet competition \cite{krizhevsky2012imagenet}.\footnote{The ImageNet Large Scale Visual Recognition Challenge (ILSVRC) released a large visual database to evaluate algorithms for classifying and detecting objects and scenes every year since 2010 \cite{imagenet_cvpr09, russakovsky2015imagenet}.}  But image classification was just the first of these benchmarks to fall.  Shortly thereafter, deep learning systems also won at object detection \cite{felzenszwalb2008discriminatively, sharma2020comprehensive,wu2020recent}, named-entity recognition \cite{li2020survey}, machine translation \cite{jean2014using, luong2014addressing, zhang2015deep}, question answering\cite{huang2020recent}, and speech recognition \cite{hannun2014deep, amodei2016deep}.
\footnotetext{The range of hardware performance values indicates the difference between SPECInt values for the lowest core counts and SPECIntRate values for the highest core counts.}

The introduction of GPU-based (and later ASIC-based) deep learning led to widespread adoption of these systems.  But the amount of computing power used in the largest cutting-edge systems grew even faster, at approximately $10\times$ per year from $2012$ to $2019$ \cite{openai2018}. This rapid increase in computing burden far outpaced the $\approx35\times$ total improvement from moving to GPUs, the meager improvements from the last vestiges of Moore's Law \cite{thompson2021decline}, or the improvements in neural network training efficiency \cite{openai2020}.  Instead, much of the increase came from a much-less-economically-attractive source: running models for more time on more machines.  For example, in $2012$ AlexNet trained using 2 GPUs for 5-6 days\cite{krizhevsky2012imagenet}, in $2017$ ResNeXt-101 \cite{xie2017aggregated} was trained with 8 GPUs for over 10 days, and in $2020$ Meta Pseudo Labels (EfficientNet-L2) was trained with $2048$ TPU cores for 11 days\cite{pham2020meta}.  Another extreme example is the machine translation system, ``Evolved Transformer'',  which used more than $2$ million GPU hours and cost millions of dollars to run\cite{So2019evolved, wu2020lite}.\footnote{Some have argued that neural architecture search costs should not be included in training costs because they help reduce inference costs \cite{jeff2022sustainable}. While we agree with the latter point, it doesn't change that this search is part of the initial training cost and thus we include it in our calculations.}  Scaling deep learning computation by scaling up hardware hours or number of chips is problematic in the longer-term because it implies that costs scale at roughly the same rate as increases in computing power \cite{openai2018}, which (as we show) will quickly make it unsustainable.

\subsection{Present}\label{sec:present}

To examine deep learning's dependence on computation, we conducted an extensive review of \textit{all} the research papers we could find that covered the domains of image classification (ImageNet benchmark, CIFAR-10, CIFAR-100), object detection (MS COCO), question answering (SQuAD 1.1), named-entity recognition (COLLN 2003), machine translation (WMT 2014), speech recognition (ASR SWB Hub500), face detection (WIDER Face Hard), image generation (CIFAR-10), and pose estimation (MPII Human Pose).  We limit our analysis to benchmarks with exact right and wrong answers, so that error rates can be precise.  For this reason, our analysis excludes generative models in language and audio.  Using established benchmarks to measure progress in these areas is important because it ensures a common baseline for comparison (which is a significant problem in parts of the deep learning literature\cite{10.1145/3298689.3347058}).  We source these deep learning papers from the arXiv repository as well as other benchmark sources (see Appendix  \ref{supp:method} for more information on the data gathering process).  

In total, we gathered $1{,}526$ deep learning papers, for which we did a detailed \textit{manual} review for their performance and computation burden data.  Unfortunately, as is well known in deep learning, few papers report the details of the amount of computation they used \cite{ethayarajh2021utility, dodge2019show}, reflecting the field's historical focus on accuracy at the expense of other measures of progress \cite{martinezplumed2018accounting}.  Most papers do not report \textit{any} details of the computational requirements for their models, and in many others only limited computational information is provided. Table \ref{tab:data_gathering} summarizes our data. Because there is insufficient computational data for other benchmarks, we limit our analysis to ImageNet, MC COCO, SQuAD 1.1, COLLN 2003, and WMT 2014 (EN-FR) and WMT 2014 (EN-DE).

\begin{table*}[t]
\centering
\caption{Deep learning benchmark data}
\label{tab:data_gathering}
\resizebox{\textwidth}{!}{%
\begin{tabular}{|ccccc|c|c|ccc|}
\hline
\multicolumn{1}{|c|}{{\cellcolor[HTML]{EFEFEF} }} &
  \multicolumn{1}{c|}{{\cellcolor[HTML]{EFEFEF} }} &
  \multicolumn{1}{c|}{{\cellcolor[HTML]{EFEFEF} }} &
  \multicolumn{1}{c|}{{\cellcolor[HTML]{EFEFEF} }} &
  {\cellcolor[HTML]{EFEFEF} } &
  {\cellcolor[HTML]{EFEFEF} } &
  {\cellcolor[HTML]{EFEFEF} } &
  \multicolumn{3}{c|}{{\cellcolor[HTML]{EFEFEF} }} \\
\multicolumn{1}{|c|}{{\cellcolor[HTML]{EFEFEF} }} &
  \multicolumn{1}{c|}{{\cellcolor[HTML]{EFEFEF} }} &
  \multicolumn{1}{c|}{{\cellcolor[HTML]{EFEFEF} }} &
  \multicolumn{1}{c|}{{\cellcolor[HTML]{EFEFEF} }} &
  {\cellcolor[HTML]{EFEFEF} } &
  {\cellcolor[HTML]{EFEFEF} } &
  {\cellcolor[HTML]{EFEFEF} } &
  \multicolumn{3}{c|}{{\cellcolor[HTML]{EFEFEF} }} \\
\multicolumn{1}{|c|}{{\cellcolor[HTML]{EFEFEF} }} &
  \multicolumn{1}{c|}{{\cellcolor[HTML]{EFEFEF} }} &
  \multicolumn{1}{c|}{{\cellcolor[HTML]{EFEFEF} }} &
  \multicolumn{1}{c|}{{\cellcolor[HTML]{EFEFEF} }} &
  {\cellcolor[HTML]{EFEFEF} } &
  {\cellcolor[HTML]{EFEFEF} } &
  {\cellcolor[HTML]{EFEFEF} } &
  \multicolumn{3}{c|}{\multirow{-3}{*}{{\cellcolor[HTML]{EFEFEF} \textbf{\# Models Including}}}} \\ \cline{8-10} 
\multicolumn{1}{|c|}{{\cellcolor[HTML]{EFEFEF} }} &
  \multicolumn{1}{c|}{{\cellcolor[HTML]{EFEFEF} }} &
  \multicolumn{1}{c|}{{\cellcolor[HTML]{EFEFEF} }} &
  \multicolumn{1}{c|}{{\cellcolor[HTML]{EFEFEF} }} &
  {\cellcolor[HTML]{EFEFEF} } &
  {\cellcolor[HTML]{EFEFEF} } &
  {\cellcolor[HTML]{EFEFEF} } &
  \multicolumn{1}{c|}{{\cellcolor[HTML]{EFEFEF} }} &
  \multicolumn{1}{c|}{{\cellcolor[HTML]{EFEFEF} }} &
  {\cellcolor[HTML]{EFEFEF} } \\
\multicolumn{1}{|c|}{{\cellcolor[HTML]{EFEFEF} }} &
  \multicolumn{1}{c|}{{\cellcolor[HTML]{EFEFEF} }} &
  \multicolumn{1}{c|}{{\cellcolor[HTML]{EFEFEF} }} &
  \multicolumn{1}{c|}{{\cellcolor[HTML]{EFEFEF} }} &
  {\cellcolor[HTML]{EFEFEF} } &
  {\cellcolor[HTML]{EFEFEF} } &
  {\cellcolor[HTML]{EFEFEF} } &
  \multicolumn{1}{c|}{{\cellcolor[HTML]{EFEFEF} }} &
  \multicolumn{1}{c|}{{\cellcolor[HTML]{EFEFEF} }} &
  {\cellcolor[HTML]{EFEFEF} } \\
\multicolumn{1}{|c|}{\multirow{-6}{*}{{\cellcolor[HTML]{EFEFEF} \textbf{Domain}}}} &
  \multicolumn{1}{c|}{\multirow{-6}{*}{{\cellcolor[HTML]{EFEFEF} \textbf{Task}}}} &
  \multicolumn{1}{c|}{\multirow{-6}{*}{{\cellcolor[HTML]{EFEFEF} \textbf{Benchmark}}}} &
  \multicolumn{1}{c|}{\multirow{-6}{*}{{\cellcolor[HTML]{EFEFEF} \textbf{\begin{tabular}[c]{@{}c@{}}Evaluation\\ Criteria\end{tabular}}}}} &
  \multirow{-6}{*}{{\cellcolor[HTML]{EFEFEF} \textbf{State-of-the-Art}}} &
  \multirow{-6}{*}{{\cellcolor[HTML]{EFEFEF} \textbf{\begin{tabular}[c]{@{}c@{}}\# Papers\\ Found\end{tabular}}}} &
  \multirow{-6}{*}{{\cellcolor[HTML]{EFEFEF} \textbf{\# Models}}} &
  \multicolumn{1}{c|}{\multirow{-3}{*}{{\cellcolor[HTML]{EFEFEF} \textbf{\begin{tabular}[c]{@{}c@{}}Hardware\\ Burden\end{tabular}}}}} &
  \multicolumn{1}{c|}{\multirow{-3}{*}{{\cellcolor[HTML]{EFEFEF} \textbf{\begin{tabular}[c]{@{}c@{}}Network\\ Operations\end{tabular}}}}} &
  \multirow{-3}{*}{{\cellcolor[HTML]{EFEFEF} \textbf{\begin{tabular}[c]{@{}c@{}}Hardware Burden \& \\ Network Operations\end{tabular}}}} \\ \hline \hline
\multicolumn{1}{|c|}{{\color[HTML]{000000} }} &
  \multicolumn{1}{c|}{\cellcolor[HTML]{6c757d}{\color[HTML]{FFFFFF} }} &
  \multicolumn{1}{c|}{\cellcolor[HTML]{6c757d}{\color[HTML]{FFFFFF} }} &
  \multicolumn{1}{c|}{\cellcolor[HTML]{6c757d}{\color[HTML]{FFFFFF} }} &
  \cellcolor[HTML]{6c757d}{\color[HTML]{FFFFFF} } &
  \cellcolor[HTML]{6c757d}{\color[HTML]{FFFFFF} } &
  \cellcolor[HTML]{6c757d}{\color[HTML]{FFFFFF} } &
  \multicolumn{1}{c|}{\cellcolor[HTML]{6c757d}{\color[HTML]{FFFFFF} }} &
  \multicolumn{1}{c|}{\cellcolor[HTML]{6c757d}{\color[HTML]{FFFFFF} }} &
  \cellcolor[HTML]{6c757d}{\color[HTML]{FFFFFF} } \\
\multicolumn{1}{|c|}{{\color[HTML]{000000} }} &
  \multicolumn{1}{c|}{\cellcolor[HTML]{6c757d}{\color[HTML]{FFFFFF} }} &
  \multicolumn{1}{c|}{\cellcolor[HTML]{6c757d}{\color[HTML]{FFFFFF} }} &
  \multicolumn{1}{c|}{\cellcolor[HTML]{6c757d}{\color[HTML]{FFFFFF} }} &
  \cellcolor[HTML]{6c757d}{\color[HTML]{FFFFFF} } &
  \cellcolor[HTML]{6c757d}{\color[HTML]{FFFFFF} } &
  \cellcolor[HTML]{6c757d}{\color[HTML]{FFFFFF} } &
  \multicolumn{1}{c|}{\cellcolor[HTML]{6c757d}{\color[HTML]{FFFFFF} }} &
  \multicolumn{1}{c|}{\cellcolor[HTML]{6c757d}{\color[HTML]{FFFFFF} }} &
  \cellcolor[HTML]{6c757d}{\color[HTML]{FFFFFF} } \\
\multicolumn{1}{|c|}{{\color[HTML]{000000} }} &
  \multicolumn{1}{c|}{\multirow{-3}{*}{\cellcolor[HTML]{6c757d}{\color[HTML]{FFFFFF} Image Classification}}} &
  \multicolumn{1}{c|}{\multirow{-3}{*}{\cellcolor[HTML]{6c757d}{\color[HTML]{FFFFFF} ImageNet}}} &
  \multicolumn{1}{c|}{\multirow{-3}{*}{\cellcolor[HTML]{6c757d}{\color[HTML]{FFFFFF} Top-1 score}}} &
  \multirow{-3}{*}{\cellcolor[HTML]{6c757d}{\color[HTML]{FFFFFF} \begin{tabular}[c]{@{}c@{}}CoCa\\ (Top 1: 91.00)\end{tabular}}} &
  \multirow{-3}{*}{\cellcolor[HTML]{6c757d}{\color[HTML]{FFFFFF} 309}} &
  \multirow{-3}{*}{\cellcolor[HTML]{6c757d}{\color[HTML]{FFFFFF} 208}} &
  \multicolumn{1}{c|}{\multirow{-3}{*}{\cellcolor[HTML]{6c757d}{\color[HTML]{FFFFFF} 36}}} &
  \multicolumn{1}{c|}{\multirow{-3}{*}{\cellcolor[HTML]{6c757d}{\color[HTML]{FFFFFF} 80}}} &
  \multirow{-3}{*}{\cellcolor[HTML]{6c757d}{\color[HTML]{FFFFFF} 11}} \\ \cline{2-10} 
\multicolumn{1}{|c|}{{\color[HTML]{000000} }} &
  \multicolumn{1}{c|}{\cellcolor[HTML]{6c757d}{\color[HTML]{FFFFFF} }} &
  \multicolumn{1}{c|}{\cellcolor[HTML]{6c757d}{\color[HTML]{FFFFFF} }} &
  \multicolumn{1}{c|}{\cellcolor[HTML]{6c757d}{\color[HTML]{FFFFFF} }} &
  \cellcolor[HTML]{6c757d}{\color[HTML]{FFFFFF} } &
  \cellcolor[HTML]{6c757d}{\color[HTML]{FFFFFF} } &
  \cellcolor[HTML]{6c757d}{\color[HTML]{FFFFFF} } &
  \multicolumn{1}{c|}{\cellcolor[HTML]{6c757d}{\color[HTML]{FFFFFF} }} &
  \multicolumn{1}{c|}{\cellcolor[HTML]{6c757d}{\color[HTML]{FFFFFF} }} &
  \cellcolor[HTML]{6c757d}{\color[HTML]{FFFFFF} } \\
\multicolumn{1}{|c|}{{\color[HTML]{000000} }} &
  \multicolumn{1}{c|}{\cellcolor[HTML]{6c757d}{\color[HTML]{FFFFFF} }} &
  \multicolumn{1}{c|}{\cellcolor[HTML]{6c757d}{\color[HTML]{FFFFFF} }} &
  \multicolumn{1}{c|}{\cellcolor[HTML]{6c757d}{\color[HTML]{FFFFFF} }} &
  \cellcolor[HTML]{6c757d}{\color[HTML]{FFFFFF} } &
  \cellcolor[HTML]{6c757d}{\color[HTML]{FFFFFF} } &
  \cellcolor[HTML]{6c757d}{\color[HTML]{FFFFFF} } &
  \multicolumn{1}{c|}{\cellcolor[HTML]{6c757d}{\color[HTML]{FFFFFF} }} &
  \multicolumn{1}{c|}{\cellcolor[HTML]{6c757d}{\color[HTML]{FFFFFF} }} &
  \cellcolor[HTML]{6c757d}{\color[HTML]{FFFFFF} } \\
\multicolumn{1}{|c|}{{\color[HTML]{000000} }} &
  \multicolumn{1}{c|}{\multirow{-3}{*}{\cellcolor[HTML]{6c757d}{\color[HTML]{FFFFFF} Object Detection}}} &
  \multicolumn{1}{c|}{\multirow{-3}{*}{\cellcolor[HTML]{6c757d}{\color[HTML]{FFFFFF} MS COCO}}} &
  \multicolumn{1}{c|}{\multirow{-3}{*}{\cellcolor[HTML]{6c757d}{\color[HTML]{FFFFFF} BOX AP}}} &
  \multirow{-3}{*}{\cellcolor[HTML]{6c757d}{\color[HTML]{FFFFFF} \begin{tabular}[c]{@{}c@{}}DINO Swin-L\\ (BOX AP: 63.3)\end{tabular}}} &
  \multirow{-3}{*}{\cellcolor[HTML]{6c757d}{\color[HTML]{FFFFFF} 277}} &
  \multirow{-3}{*}{\cellcolor[HTML]{6c757d}{\color[HTML]{FFFFFF} 116}} &
  \multicolumn{1}{c|}{\multirow{-3}{*}{\cellcolor[HTML]{6c757d}{\color[HTML]{FFFFFF} 13}}} &
  \multicolumn{1}{c|}{\multirow{-3}{*}{\cellcolor[HTML]{6c757d}{\color[HTML]{FFFFFF} 7}}} &
  \multirow{-3}{*}{\cellcolor[HTML]{6c757d}{\color[HTML]{FFFFFF} 0}} \\ \cline{2-10} 
\multicolumn{1}{|c|}{{\color[HTML]{000000} }} &
  \multicolumn{1}{c|}{{\color[HTML]{000000} }} &
  \multicolumn{1}{c|}{{\color[HTML]{000000} }} &
  \multicolumn{1}{c|}{{\color[HTML]{000000} }} &
  {\color[HTML]{000000} } &
  {\color[HTML]{000000} } &
  {\color[HTML]{000000} } &
  \multicolumn{1}{c|}{{\color[HTML]{000000} }} &
  \multicolumn{1}{c|}{{\color[HTML]{000000} }} &
  {\color[HTML]{000000} } \\
\multicolumn{1}{|c|}{{\color[HTML]{000000} }} &
  \multicolumn{1}{c|}{{\color[HTML]{000000} }} &
  \multicolumn{1}{c|}{{\color[HTML]{000000} }} &
  \multicolumn{1}{c|}{{\color[HTML]{000000} }} &
  {\color[HTML]{000000} } &
  {\color[HTML]{000000} } &
  {\color[HTML]{000000} } &
  \multicolumn{1}{c|}{{\color[HTML]{000000} }} &
  \multicolumn{1}{c|}{{\color[HTML]{000000} }} &
  {\color[HTML]{000000} } \\
\multicolumn{1}{|c|}{{\color[HTML]{000000} }} &
  \multicolumn{1}{c|}{\multirow{-3}{*}{{\color[HTML]{000000} Image Classification}}} &
  \multicolumn{1}{c|}{\multirow{-3}{*}{{\color[HTML]{000000} CIFAR-10}}} &
  \multicolumn{1}{c|}{\multirow{-3}{*}{{\color[HTML]{000000} Percentage Correct}}} &
  \multirow{-3}{*}{{\color[HTML]{000000} \begin{tabular}[c]{@{}c@{}}ViT-H/14\\ (Percentage Correct: 99.5)\end{tabular}}} &
  \multirow{-3}{*}{{\color[HTML]{000000} 78}} &
  \multirow{-3}{*}{{\color[HTML]{000000} 94}} &
  \multicolumn{1}{c|}{\multirow{-3}{*}{{\color[HTML]{000000} 1}}} &
  \multicolumn{1}{c|}{\multirow{-3}{*}{{\color[HTML]{000000} 0}}} &
  \multirow{-3}{*}{{\color[HTML]{000000} 0}} \\ \cline{2-10} 
\multicolumn{1}{|c|}{{\color[HTML]{000000} }} &
  \multicolumn{1}{c|}{{\color[HTML]{000000} }} &
  \multicolumn{1}{c|}{{\color[HTML]{000000} }} &
  \multicolumn{1}{c|}{{\color[HTML]{000000} }} &
  {\color[HTML]{000000} } &
  {\color[HTML]{000000} } &
  {\color[HTML]{000000} } &
  \multicolumn{1}{c|}{{\color[HTML]{000000} }} &
  \multicolumn{1}{c|}{{\color[HTML]{000000} }} &
  {\color[HTML]{000000} } \\
\multicolumn{1}{|c|}{{\color[HTML]{000000} }} &
  \multicolumn{1}{c|}{{\color[HTML]{000000} }} &
  \multicolumn{1}{c|}{{\color[HTML]{000000} }} &
  \multicolumn{1}{c|}{{\color[HTML]{000000} }} &
  {\color[HTML]{000000} } &
  {\color[HTML]{000000} } &
  {\color[HTML]{000000} } &
  \multicolumn{1}{c|}{{\color[HTML]{000000} }} &
  \multicolumn{1}{c|}{{\color[HTML]{000000} }} &
  {\color[HTML]{000000} } \\
\multicolumn{1}{|c|}{{\color[HTML]{000000} }} &
  \multicolumn{1}{c|}{\multirow{-3}{*}{{\color[HTML]{000000} Image Classification}}} &
  \multicolumn{1}{c|}{\multirow{-3}{*}{{\color[HTML]{000000} CIFAR-100}}} &
  \multicolumn{1}{c|}{\multirow{-3}{*}{{\color[HTML]{000000} Percentage Correct}}} &
  \multirow{-3}{*}{{\color[HTML]{000000} \begin{tabular}[c]{@{}c@{}}EffNet-L2 SAM\\ (Percentage Correct: 96.08)\end{tabular}}} &
  \multirow{-3}{*}{{\color[HTML]{000000} 70}} &
  \multirow{-3}{*}{{\color[HTML]{000000} 71}} &
  \multicolumn{1}{c|}{\multirow{-3}{*}{{\color[HTML]{000000} 0}}} &
  \multicolumn{1}{c|}{\multirow{-3}{*}{{\color[HTML]{000000} 0}}} &
  \multirow{-3}{*}{{\color[HTML]{000000} 0}} \\ \cline{2-10} 
\multicolumn{1}{|c|}{{\color[HTML]{000000} }} &
  \multicolumn{1}{c|}{{\color[HTML]{000000} }} &
  \multicolumn{1}{c|}{{\color[HTML]{000000} }} &
  \multicolumn{1}{c|}{{\color[HTML]{000000} }} &
  {\color[HTML]{000000} } &
  {\color[HTML]{000000} } &
  {\color[HTML]{000000} } &
  \multicolumn{1}{c|}{{\color[HTML]{000000} }} &
  \multicolumn{1}{c|}{{\color[HTML]{000000} }} &
  {\color[HTML]{000000} } \\
\multicolumn{1}{|c|}{{\color[HTML]{000000} }} &
  \multicolumn{1}{c|}{{\color[HTML]{000000} }} &
  \multicolumn{1}{c|}{{\color[HTML]{000000} }} &
  \multicolumn{1}{c|}{{\color[HTML]{000000} }} &
  {\color[HTML]{000000} } &
  {\color[HTML]{000000} } &
  {\color[HTML]{000000} } &
  \multicolumn{1}{c|}{{\color[HTML]{000000} }} &
  \multicolumn{1}{c|}{{\color[HTML]{000000} }} &
  {\color[HTML]{000000} } \\
\multicolumn{1}{|c|}{{\color[HTML]{000000} }} &
  \multicolumn{1}{c|}{\multirow{-3}{*}{{\color[HTML]{000000} Face Detection}}} &
  \multicolumn{1}{c|}{\multirow{-3}{*}{{\color[HTML]{000000} WIDER Face (Hard)}}} &
  \multicolumn{1}{c|}{\multirow{-3}{*}{{\color[HTML]{000000} AP}}} &
  \multirow{-3}{*}{{\color[HTML]{000000} \begin{tabular}[c]{@{}c@{}}TinaFace\\ (AP: 0.924)\end{tabular}}} &
  \multirow{-3}{*}{{\color[HTML]{000000} 21}} &
  \multirow{-3}{*}{{\color[HTML]{000000} 21}} &
  \multicolumn{1}{c|}{\multirow{-3}{*}{{\color[HTML]{000000} 0}}} &
  \multicolumn{1}{c|}{\multirow{-3}{*}{{\color[HTML]{000000} 0}}} &
  \multirow{-3}{*}{{\color[HTML]{000000} 0}} \\ \cline{2-10} 
\multicolumn{1}{|c|}{{\color[HTML]{000000} }} &
  \multicolumn{1}{c|}{{\color[HTML]{000000} }} &
  \multicolumn{1}{c|}{{\color[HTML]{000000} }} &
  \multicolumn{1}{c|}{{\color[HTML]{000000} }} &
  {\color[HTML]{000000} } &
  {\color[HTML]{000000} } &
  {\color[HTML]{000000} } &
  \multicolumn{1}{c|}{{\color[HTML]{000000} }} &
  \multicolumn{1}{c|}{{\color[HTML]{000000} }} &
  {\color[HTML]{000000} } \\
\multicolumn{1}{|c|}{{\color[HTML]{000000} }} &
  \multicolumn{1}{c|}{{\color[HTML]{000000} }} &
  \multicolumn{1}{c|}{{\color[HTML]{000000} }} &
  \multicolumn{1}{c|}{{\color[HTML]{000000} }} &
  {\color[HTML]{000000} } &
  {\color[HTML]{000000} } &
  {\color[HTML]{000000} } &
  \multicolumn{1}{c|}{{\color[HTML]{000000} }} &
  \multicolumn{1}{c|}{{\color[HTML]{000000} }} &
  {\color[HTML]{000000} } \\
\multicolumn{1}{|c|}{{\color[HTML]{000000} }} &
  \multicolumn{1}{c|}{\multirow{-3}{*}{{\color[HTML]{000000} Image Generation}}} &
  \multicolumn{1}{c|}{\multirow{-3}{*}{{\color[HTML]{000000} CIFAR-10}}} &
  \multicolumn{1}{c|}{\multirow{-3}{*}{{\color[HTML]{000000} FID}}} &
  \multirow{-3}{*}{{\color[HTML]{000000} \begin{tabular}[c]{@{}c@{}}LSGM\\ (FID: 2.1)\end{tabular}}} &
  \multirow{-3}{*}{{\color[HTML]{000000} 35}} &
  \multirow{-3}{*}{{\color[HTML]{000000} 35}} &
  \multicolumn{1}{c|}{\multirow{-3}{*}{{\color[HTML]{000000} 4}}} &
  \multicolumn{1}{c|}{\multirow{-3}{*}{{\color[HTML]{000000} 0}}} &
  \multirow{-3}{*}{{\color[HTML]{000000} 0}} \\ \cline{2-10} 
\multicolumn{1}{|c|}{{\color[HTML]{000000} }} &
  \multicolumn{1}{c|}{{\color[HTML]{000000} }} &
  \multicolumn{1}{c|}{{\color[HTML]{000000} }} &
  \multicolumn{1}{c|}{{\color[HTML]{000000} }} &
  {\color[HTML]{000000} } &
  {\color[HTML]{000000} } &
  {\color[HTML]{000000} } &
  \multicolumn{1}{c|}{{\color[HTML]{000000} }} &
  \multicolumn{1}{c|}{{\color[HTML]{000000} }} &
  {\color[HTML]{000000} } \\
\multicolumn{1}{|c|}{{\color[HTML]{000000} }} &
  \multicolumn{1}{c|}{{\color[HTML]{000000} }} &
  \multicolumn{1}{c|}{{\color[HTML]{000000} }} &
  \multicolumn{1}{c|}{{\color[HTML]{000000} }} &
  {\color[HTML]{000000} } &
  {\color[HTML]{000000} } &
  {\color[HTML]{000000} } &
  \multicolumn{1}{c|}{{\color[HTML]{000000} }} &
  \multicolumn{1}{c|}{{\color[HTML]{000000} }} &
  {\color[HTML]{000000} } \\
\multicolumn{1}{|c|}{\multirow{-21}{*}{{\color[HTML]{000000} Computer Vision}}} &
  \multicolumn{1}{c|}{\multirow{-3}{*}{{\color[HTML]{000000} Pose Estimation}}} &
  \multicolumn{1}{c|}{\multirow{-3}{*}{{\color[HTML]{000000} MPII Human Pose}}} &
  \multicolumn{1}{c|}{\multirow{-3}{*}{{\color[HTML]{000000} PCKh-0.5}}} &
  \multirow{-3}{*}{{\color[HTML]{000000} \begin{tabular}[c]{@{}c@{}}Soft-gated Skip Connections\\ (PCKh-0.5: 94.1)\end{tabular}}} &
  \multirow{-3}{*}{{\color[HTML]{000000} 30}} &
  \multirow{-3}{*}{{\color[HTML]{000000} 30}} &
  \multicolumn{1}{c|}{\multirow{-3}{*}{{\color[HTML]{000000} 3}}} &
  \multicolumn{1}{c|}{\multirow{-3}{*}{{\color[HTML]{000000} 4}}} &
  \multirow{-3}{*}{{\color[HTML]{000000} 0}} \\ \hline
\multicolumn{1}{|c|}{{\color[HTML]{000000} }} &
  \multicolumn{1}{c|}{\cellcolor[HTML]{6c757d}{\color[HTML]{FFFFFF} }} &
  \multicolumn{1}{c|}{\cellcolor[HTML]{6c757d}{\color[HTML]{FFFFFF} }} &
  \multicolumn{1}{c|}{\cellcolor[HTML]{6c757d}{\color[HTML]{FFFFFF} }} &
  \cellcolor[HTML]{6c757d}{\color[HTML]{FFFFFF} } &
  \cellcolor[HTML]{6c757d}{\color[HTML]{FFFFFF} } &
  \cellcolor[HTML]{6c757d}{\color[HTML]{FFFFFF} } &
  \multicolumn{1}{c|}{\cellcolor[HTML]{6c757d}{\color[HTML]{FFFFFF} }} &
  \multicolumn{1}{c|}{\cellcolor[HTML]{6c757d}{\color[HTML]{FFFFFF} }} &
  \cellcolor[HTML]{6c757d}{\color[HTML]{FFFFFF} } \\
\multicolumn{1}{|c|}{{\color[HTML]{000000} }} &
  \multicolumn{1}{c|}{\cellcolor[HTML]{6c757d}{\color[HTML]{FFFFFF} }} &
  \multicolumn{1}{c|}{\cellcolor[HTML]{6c757d}{\color[HTML]{FFFFFF} }} &
  \multicolumn{1}{c|}{\cellcolor[HTML]{6c757d}{\color[HTML]{FFFFFF} }} &
  \cellcolor[HTML]{6c757d}{\color[HTML]{FFFFFF} } &
  \cellcolor[HTML]{6c757d}{\color[HTML]{FFFFFF} } &
  \cellcolor[HTML]{6c757d}{\color[HTML]{FFFFFF} } &
  \multicolumn{1}{c|}{\cellcolor[HTML]{6c757d}{\color[HTML]{FFFFFF} }} &
  \multicolumn{1}{c|}{\cellcolor[HTML]{6c757d}{\color[HTML]{FFFFFF} }} &
  \cellcolor[HTML]{6c757d}{\color[HTML]{FFFFFF} } \\
\multicolumn{1}{|c|}{{\color[HTML]{000000} }} &
  \multicolumn{1}{c|}{\multirow{-3}{*}{\cellcolor[HTML]{6c757d}{\color[HTML]{FFFFFF} Question Answering}}} &
  \multicolumn{1}{c|}{\multirow{-3}{*}{\cellcolor[HTML]{6c757d}{\color[HTML]{FFFFFF} SQuAD 1.1}}} &
  \multicolumn{1}{c|}{\multirow{-3}{*}{\cellcolor[HTML]{6c757d}{\color[HTML]{FFFFFF} EM}}} &
  \multirow{-3}{*}{\cellcolor[HTML]{6c757d}{\color[HTML]{FFFFFF} \begin{tabular}[c]{@{}c@{}}ANNA\\ (EM: 90.6)\end{tabular}}} &
  \multirow{-3}{*}{\cellcolor[HTML]{6c757d}{\color[HTML]{FFFFFF} 147}} &
  \multirow{-3}{*}{\cellcolor[HTML]{6c757d}{\color[HTML]{FFFFFF} 41}} &
  \multicolumn{1}{c|}{\multirow{-3}{*}{\cellcolor[HTML]{6c757d}{\color[HTML]{FFFFFF} 12}}} &
  \multicolumn{1}{c|}{\multirow{-3}{*}{\cellcolor[HTML]{6c757d}{\color[HTML]{FFFFFF} 0}}} &
  \multirow{-3}{*}{\cellcolor[HTML]{6c757d}{\color[HTML]{FFFFFF} 0}} \\ \cline{2-10} 
\multicolumn{1}{|c|}{{\color[HTML]{000000} }} &
  \multicolumn{1}{c|}{\cellcolor[HTML]{6c757d}{\color[HTML]{FFFFFF} }} &
  \multicolumn{1}{c|}{\cellcolor[HTML]{6c757d}{\color[HTML]{FFFFFF} }} &
  \multicolumn{1}{c|}{\cellcolor[HTML]{6c757d}{\color[HTML]{FFFFFF} }} &
  \cellcolor[HTML]{6c757d}{\color[HTML]{FFFFFF} } &
  \cellcolor[HTML]{6c757d}{\color[HTML]{FFFFFF} } &
  \cellcolor[HTML]{6c757d}{\color[HTML]{FFFFFF} } &
  \multicolumn{1}{c|}{\cellcolor[HTML]{6c757d}{\color[HTML]{FFFFFF} }} &
  \multicolumn{1}{c|}{\cellcolor[HTML]{6c757d}{\color[HTML]{FFFFFF} }} &
  \cellcolor[HTML]{6c757d}{\color[HTML]{FFFFFF} } \\
\multicolumn{1}{|c|}{{\color[HTML]{000000} }} &
  \multicolumn{1}{c|}{\cellcolor[HTML]{6c757d}{\color[HTML]{FFFFFF} }} &
  \multicolumn{1}{c|}{\cellcolor[HTML]{6c757d}{\color[HTML]{FFFFFF} }} &
  \multicolumn{1}{c|}{\cellcolor[HTML]{6c757d}{\color[HTML]{FFFFFF} }} &
  \cellcolor[HTML]{6c757d}{\color[HTML]{FFFFFF} } &
  \cellcolor[HTML]{6c757d}{\color[HTML]{FFFFFF} } &
  \cellcolor[HTML]{6c757d}{\color[HTML]{FFFFFF} } &
  \multicolumn{1}{c|}{\cellcolor[HTML]{6c757d}{\color[HTML]{FFFFFF} }} &
  \multicolumn{1}{c|}{\cellcolor[HTML]{6c757d}{\color[HTML]{FFFFFF} }} &
  \cellcolor[HTML]{6c757d}{\color[HTML]{FFFFFF} } \\
\multicolumn{1}{|c|}{{\color[HTML]{000000} }} &
  \multicolumn{1}{c|}{\multirow{-3}{*}{\cellcolor[HTML]{6c757d}{\color[HTML]{FFFFFF} Named Entity   Recognition}}} &
  \multicolumn{1}{c|}{\multirow{-3}{*}{\cellcolor[HTML]{6c757d}{\color[HTML]{FFFFFF} CoNLL 2003}}} &
  \multicolumn{1}{c|}{\multirow{-3}{*}{\cellcolor[HTML]{6c757d}{\color[HTML]{FFFFFF} F1-score}}} &
  \multirow{-3}{*}{\cellcolor[HTML]{6c757d}{\color[HTML]{FFFFFF} \begin{tabular}[c]{@{}c@{}}ACE + document-context\\ (F1-score: 94.6)\end{tabular}}} &
  \multirow{-3}{*}{\cellcolor[HTML]{6c757d}{\color[HTML]{FFFFFF} 247}} &
  \multirow{-3}{*}{\cellcolor[HTML]{6c757d}{\color[HTML]{FFFFFF} 54}} &
  \multicolumn{1}{c|}{\multirow{-3}{*}{\cellcolor[HTML]{6c757d}{\color[HTML]{FFFFFF} 12}}} &
  \multicolumn{1}{c|}{\multirow{-3}{*}{\cellcolor[HTML]{6c757d}{\color[HTML]{FFFFFF} 0}}} &
  \multirow{-3}{*}{\cellcolor[HTML]{6c757d}{\color[HTML]{FFFFFF} 0}} \\ \cline{2-10} 
\multicolumn{1}{|c|}{{\color[HTML]{000000} }} &
  \multicolumn{1}{c|}{\cellcolor[HTML]{6c757d}{\color[HTML]{FFFFFF} }} &
  \multicolumn{1}{c|}{\cellcolor[HTML]{6c757d}{\color[HTML]{FFFFFF} }} &
  \multicolumn{1}{c|}{\cellcolor[HTML]{6c757d}{\color[HTML]{FFFFFF} }} &
  \cellcolor[HTML]{6c757d}{\color[HTML]{FFFFFF} } &
  \cellcolor[HTML]{6c757d}{\color[HTML]{FFFFFF} } &
  \cellcolor[HTML]{6c757d}{\color[HTML]{FFFFFF} } &
  \multicolumn{1}{c|}{\cellcolor[HTML]{6c757d}{\color[HTML]{FFFFFF} }} &
  \multicolumn{1}{c|}{\cellcolor[HTML]{6c757d}{\color[HTML]{FFFFFF} }} &
  \cellcolor[HTML]{6c757d}{\color[HTML]{FFFFFF} } \\
\multicolumn{1}{|c|}{{\color[HTML]{000000} }} &
  \multicolumn{1}{c|}{\cellcolor[HTML]{6c757d}{\color[HTML]{FFFFFF} }} &
  \multicolumn{1}{c|}{\cellcolor[HTML]{6c757d}{\color[HTML]{FFFFFF} }} &
  \multicolumn{1}{c|}{\cellcolor[HTML]{6c757d}{\color[HTML]{FFFFFF} }} &
  \cellcolor[HTML]{6c757d}{\color[HTML]{FFFFFF} } &
  \cellcolor[HTML]{6c757d}{\color[HTML]{FFFFFF} } &
  \cellcolor[HTML]{6c757d}{\color[HTML]{FFFFFF} } &
  \multicolumn{1}{c|}{\cellcolor[HTML]{6c757d}{\color[HTML]{FFFFFF} }} &
  \multicolumn{1}{c|}{\cellcolor[HTML]{6c757d}{\color[HTML]{FFFFFF} }} &
  \cellcolor[HTML]{6c757d}{\color[HTML]{FFFFFF} } \\
\multicolumn{1}{|c|}{{\color[HTML]{000000} }} &
  \multicolumn{1}{c|}{\multirow{-3}{*}{\cellcolor[HTML]{6c757d}{\color[HTML]{FFFFFF} Machine Translation}}} &
  \multicolumn{1}{c|}{\multirow{-3}{*}{\cellcolor[HTML]{6c757d}{\color[HTML]{FFFFFF} WMT 2014 (EN-FR)}}} &
  \multicolumn{1}{c|}{\multirow{-3}{*}{\cellcolor[HTML]{6c757d}{\color[HTML]{FFFFFF} BLEU}}} &
  \multirow{-3}{*}{\cellcolor[HTML]{6c757d}{\color[HTML]{FFFFFF} \begin{tabular}[c]{@{}c@{}}Transformer + BT ADMIN init\\ (BLEU: 46.4)\end{tabular}}} &
  \multirow{-3}{*}{\cellcolor[HTML]{6c757d}{\color[HTML]{FFFFFF} 96}} &
  \multirow{-3}{*}{\cellcolor[HTML]{6c757d}{\color[HTML]{FFFFFF} 42}} &
  \multicolumn{1}{c|}{\multirow{-3}{*}{\cellcolor[HTML]{6c757d}{\color[HTML]{FFFFFF} 14}}} &
  \multicolumn{1}{c|}{\multirow{-3}{*}{\cellcolor[HTML]{6c757d}{\color[HTML]{FFFFFF} 0}}} &
  \multirow{-3}{*}{\cellcolor[HTML]{6c757d}{\color[HTML]{FFFFFF} 0}} \\ \cline{2-10} 
\multicolumn{1}{|c|}{{\color[HTML]{000000} }} &
  \multicolumn{1}{c|}{\cellcolor[HTML]{6c757d}{\color[HTML]{FFFFFF} }} &
  \multicolumn{1}{c|}{\cellcolor[HTML]{6c757d}{\color[HTML]{FFFFFF} }} &
  \multicolumn{1}{c|}{\cellcolor[HTML]{6c757d}{\color[HTML]{FFFFFF} }} &
  \cellcolor[HTML]{6c757d}{\color[HTML]{FFFFFF} } &
  \cellcolor[HTML]{6c757d}{\color[HTML]{FFFFFF} } &
  \cellcolor[HTML]{6c757d}{\color[HTML]{FFFFFF} } &
  \multicolumn{1}{c|}{\cellcolor[HTML]{6c757d}{\color[HTML]{FFFFFF} }} &
  \multicolumn{1}{c|}{\cellcolor[HTML]{6c757d}{\color[HTML]{FFFFFF} }} &
  \cellcolor[HTML]{6c757d}{\color[HTML]{FFFFFF} } \\
\multicolumn{1}{|c|}{{\color[HTML]{000000} }} &
  \multicolumn{1}{c|}{\cellcolor[HTML]{6c757d}{\color[HTML]{FFFFFF} }} &
  \multicolumn{1}{c|}{\cellcolor[HTML]{6c757d}{\color[HTML]{FFFFFF} }} &
  \multicolumn{1}{c|}{\cellcolor[HTML]{6c757d}{\color[HTML]{FFFFFF} }} &
  \cellcolor[HTML]{6c757d}{\color[HTML]{FFFFFF} } &
  \cellcolor[HTML]{6c757d}{\color[HTML]{FFFFFF} } &
  \cellcolor[HTML]{6c757d}{\color[HTML]{FFFFFF} } &
  \multicolumn{1}{c|}{\cellcolor[HTML]{6c757d}{\color[HTML]{FFFFFF} }} &
  \multicolumn{1}{c|}{\cellcolor[HTML]{6c757d}{\color[HTML]{FFFFFF} }} &
  \cellcolor[HTML]{6c757d}{\color[HTML]{FFFFFF} } \\
\multicolumn{1}{|c|}{\multirow{-12}{*}{{\color[HTML]{000000} Natural Language   Processing}}} &
  \multicolumn{1}{c|}{\multirow{-3}{*}{\cellcolor[HTML]{6c757d}{\color[HTML]{FFFFFF} Machine Translation}}} &
  \multicolumn{1}{c|}{\multirow{-3}{*}{\cellcolor[HTML]{6c757d}{\color[HTML]{FFFFFF} WMT 2014 (EN-DE)}}} &
  \multicolumn{1}{c|}{\multirow{-3}{*}{\cellcolor[HTML]{6c757d}{\color[HTML]{FFFFFF} BLEU}}} &
  \multirow{-3}{*}{\cellcolor[HTML]{6c757d}{\color[HTML]{FFFFFF} \begin{tabular}[c]{@{}c@{}}Transformer Cycle Rev\\ (BLEU: 35.14)\end{tabular}}} &
  \multirow{-3}{*}{\cellcolor[HTML]{6c757d}{\color[HTML]{FFFFFF} 127}} &
  \multirow{-3}{*}{\cellcolor[HTML]{6c757d}{\color[HTML]{FFFFFF} 55}} &
  \multicolumn{1}{c|}{\multirow{-3}{*}{\cellcolor[HTML]{6c757d}{\color[HTML]{FFFFFF} 12}}} &
  \multicolumn{1}{c|}{\multirow{-3}{*}{\cellcolor[HTML]{6c757d}{\color[HTML]{FFFFFF} 0}}} &
  \multirow{-3}{*}{\cellcolor[HTML]{6c757d}{\color[HTML]{FFFFFF} 0}} \\ \hline
\multicolumn{1}{|c|}{{\color[HTML]{000000} }} &
  \multicolumn{1}{c|}{{\color[HTML]{000000} }} &
  \multicolumn{1}{c|}{{\color[HTML]{000000} }} &
  \multicolumn{1}{c|}{{\color[HTML]{000000} }} &
  {\color[HTML]{000000} } &
  {\color[HTML]{000000} } &
  {\color[HTML]{000000} } &
  \multicolumn{1}{c|}{{\color[HTML]{000000} }} &
  \multicolumn{1}{c|}{{\color[HTML]{000000} }} &
  {\color[HTML]{000000} } \\
\multicolumn{1}{|c|}{{\color[HTML]{000000} }} &
  \multicolumn{1}{c|}{{\color[HTML]{000000} }} &
  \multicolumn{1}{c|}{{\color[HTML]{000000} }} &
  \multicolumn{1}{c|}{{\color[HTML]{000000} }} &
  {\color[HTML]{000000} } &
  {\color[HTML]{000000} } &
  {\color[HTML]{000000} } &
  \multicolumn{1}{c|}{{\color[HTML]{000000} }} &
  \multicolumn{1}{c|}{{\color[HTML]{000000} }} &
  {\color[HTML]{000000} } \\
\multicolumn{1}{|c|}{\multirow{-3}{*}{{\color[HTML]{000000} Speech}}} &
  \multicolumn{1}{c|}{\multirow{-3}{*}{{\color[HTML]{000000} Speech Recognition}}} &
  \multicolumn{1}{c|}{\multirow{-3}{*}{{\color[HTML]{000000} Switchboard + Hub 500}}} &
  \multicolumn{1}{c|}{\multirow{-3}{*}{{\color[HTML]{000000} Percentage Error}}} &
  \multirow{-3}{*}{{\color[HTML]{000000} \begin{tabular}[c]{@{}c@{}}IBM LSTM + Conformer encoder-decoder\\ (Percentage Error: 4.3)\end{tabular}}} &
  \multirow{-3}{*}{{\color[HTML]{000000} 90}} &
  \multirow{-3}{*}{{\color[HTML]{000000} 18}} &
  \multicolumn{1}{c|}{\multirow{-3}{*}{{\color[HTML]{000000} 7}}} &
  \multicolumn{1}{c|}{\multirow{-3}{*}{{\color[HTML]{000000} 0}}} &
  \multirow{-3}{*}{{\color[HTML]{000000} 0}} \\ \hline
\multicolumn{5}{|c|}{{\color[HTML]{000000} }} &
  {\color[HTML]{000000} } &
  {\color[HTML]{000000} } &
  \multicolumn{1}{c|}{{\color[HTML]{000000} }} &
  \multicolumn{1}{c|}{{\color[HTML]{000000} }} &
  {\color[HTML]{000000} } \\
\multicolumn{5}{|c|}{\multirow{-2}{*}{{\color[HTML]{000000} \textbf{Total}}}} &
  \multirow{-2}{*}{{\color[HTML]{000000} \textbf{1527}}} &
  \multirow{-2}{*}{{\color[HTML]{000000} \textbf{785}}} &
  \multicolumn{1}{c|}{\multirow{-2}{*}{{\color[HTML]{000000} \textbf{114}}}} &
  \multicolumn{1}{c|}{\multirow{-2}{*}{{\color[HTML]{000000} \textbf{91}}}} &
  \multirow{-2}{*}{{\color[HTML]{000000} \textbf{11}}} \\ \hline
\end{tabular}}
\end{table*}

Reflecting the computational information available in these papers, we do separate analyses for two measures of the computational burden: (\ref{eqn:net_ops}) Network Operations\cite{openai2018, sevilla2022compute}, the number of floating point operations computed in the network\footnote{Note that, one $multiply-add$ operation is composed of two arithmetic operations (the product of two numbers and the addition of this product to an accumulator). Therefore, for cases where authors report the number of forward-pass operations in $multiply-add$ operations, we use a conversion factor equals to $2$ to convert this value to $flops$ \cite{openai2018}}, and (\ref{eqn:hw_burden}) Hardware burden, the computational capacity of the hardware used to train the model.

\begin{equation}
    \label{eqn:net_ops}
    \sum_{\substack{i \in \{pre-training,\\ training,\\ fine-tuning\}}}{Epochs_i \cdot FlopsPerNetworkPass_i \cdot NetworkPassesPerEpoch_i}
\end{equation}

\begin{equation}
    \label{eqn:hw_burden}
    \sum_{\substack{i \in \{pre-training,\\ training,\\ fine-tuning\}}}{Processors_i \cdot ComputationRate_i \cdot Time_i}
\end{equation}

We illustrate our analysis first in the area with the most data and longest history: image classification. The relevant performance metric here is the error rate for classification. As discussed in the previous section, we should expect computation to scale at least as a high order (e.g. 4th order) polynomial versus performance, which we estimate via the equation: $\log(1/error)=\alpha+\beta\cdot \log(computation)$.

Figure \ref{fig:long_term_growth} (b) shows the fall in the image classification error rate for the ImageNet dataset and its correlation to the computation used in these models.  Each data point reflects a particular deep learning model from the literature. Because this is plotted on a log-log scale, a straight line indicates a polynomial growth in computing per unit of performance -- that is, a power law.  In particular, a polynomial relationship between computation and performance of the form \(Computation = Performance^\alpha \) yields a slope of \(-\frac{1}{\alpha}\) in our plots.  Thus, our estimated slope coefficient of $-0.08$ (p-value $< 0.01$) indicates that computation used for ImageNet scales as \(\mathcal{O}(Performance^{12.5})\).  Concretely, this means that every halving of the remaining error on this problem requires $\approx2^{12.5}>5{,}000\times$ as much computation.

Taking into account the standard error on this estimate yields a 95\% confidence interval for scaling between \(\mathcal{O}(Performance^{10.6})\) and \(\mathcal{O}(Performance^{14.1})\), i.e. between $\approx1{,}500\times$ and $\approx17{,}500\times$ as much computation to halve the error.  Not only is computational power a highly statistically significant predictor of performance, but it also has substantial explanatory power, explaining 71\% of the variance in ImageNet performance.  Table \ref{tab:deep_learning_regressions}a, specification $1$, reports this regression result, alongside a series of alternate specifications that test the robustness of our finding.

\begin{table*}[!htbp] \centering 
\caption{Regression Analysis of how Deep Learning Performance depends on Computing Power Growth}
\resizebox{\textwidth}{!}{
\begin{tabular}{@{\extracolsep{5pt}}lcccccc}
\\[-1.8ex]\hline
\hline \\ \\
[-1.8ex] &  \multicolumn{1}{c}{} & \multicolumn{1}{c}{(1)} & \multicolumn{1}{c}{(2)} & \multicolumn{1}{c}{(3)} & \multicolumn{1}{c}{(4)} \\ \\

\\[-1.8ex]  & & $\log_{10}$(Top\ 1\ error) & $\log_{10}$(Top\ 1\ error) & Top\ 1\ error & $\log_{10}$(Top\ 1\ error) \\ 
\textbf{\large (a) Network Operations}
\\

\\[-1.8ex]  & & Image Classification & Image Classification & Image Classification & Image Classification \\ \\
[-1.8ex]  & &  (Imagenet) & (Imagenet) & (Imagenet) & (Imagenet) \\ \\

\\[-1.8ex]  & & OLS\ Regression & OLS\ Regression & OLS\ Regression & Quantile\ Regression\\ \\

\hline \\[-1.8ex] \\
 $log_{10}(NetworkOperations)$ & & -0.082$^{***}$ & -0.065$^{***}$ & -0.033$^{***}$ & -0.084$^{***}$ \\
  & & (0.006) & (0.005) & (0.003) & (0.005) \\ \\
 $Year$ & & & -0.022$^{***}$ & &\\
  & &  & (0.003) & & \\ \\
 $Intercept$ & & -0.629$^{***}$ & -0.476$^{***}$ & 0.231$^{***}$ & -0.676$^{***}$ \\
  &  &  (0.011) & (0.024) & (0.005) & (0.010)\\
\hline \\[-1.8ex]\\
 Observations & & 80 & 80 & 80 & 80 \\
 $R^2\ /\ pseudo\ R^2$ & & 0.712 & 0.822 & 0.622 & 0.557 \\
 Adjusted $R^2$ & & 0.708 & 0.817 & 0.617 & $-$ \\
 Residual Std. Error & & 0.051 (df =78) & 0.040 (df = 77) & 0.025 (df = 78) & $-$  \\
 F Statistic & & 192.68$^{***}$ (df = 1; 78) & 177.768$^{***}$ (df = 2; 77) & 128.163$^{***}$ (df = 1; 78) & $-$\\ \\
 \hline \\
 Implied\ Polynomial\ Scaling\ Factor & & 12.1 & 15.5 & $-$ & 11.9\\
 95\%\ Confidence\ Interval & & 10.6\ $-$\ 14.1 & 13.3\ $-$\ 18.5 & $-$ & 10.6\ $-$\ 13.5 \\ \\
 \hline
 \hline \\[-1.8ex] \\ \\
 [-1.8ex] & \multicolumn{1}{c}{} &\multicolumn{1}{c}{(5)} & \multicolumn{1}{c}{(6)} & \multicolumn{1}{c}{(7)} & \multicolumn{1}{c}{(8)} & \\ \\
 \textbf{\large (b) Hardware Burden} &
 & \multicolumn{1}{c}{$\log_{10}$ (TOP\ 1)} & \multicolumn{1}{c}{$\log_{10}$(BOX\ AP)} & \multicolumn{1}{c}{$\log_{10}$(EM)} & \multicolumn{1}{c}{$\log_{10}$(F1\ score)} \\ \\
 
   \multicolumn{1}{c}{} &
  & \multicolumn{1}{c}{Image Classification}  &  \multicolumn{1}{c}{Object Detection} &  \multicolumn{1}{c}{Question Answering} &  \multicolumn{1}{c}{Named Entity Recognition} \\ 
  \multicolumn{1}{c}{} &
  & \multicolumn{1}{c}{(ImageNet)}  &  \multicolumn{1}{c}{(MS\ COCO)} &  \multicolumn{1}{c}{(SQuAD\ 1.1)} &  \multicolumn{1}{c}{(CoNLL\ 2003)} \\ 
 \\
  \hline \\[-1.8ex] \\ 
  $log_{10} (HardwareBurden)$ & & 0.093$^{***}$ & 0.062$^{***}$ & 0.096$^{***}$ & 0.027$^{**}$ \\ 
  & & (0.006) & (0.012) & (0.012) & (0.010) \\ 
  & & & & & \\ 
 $Intercept$ & & -1.111$^{***}$ & -0.962$^{***}$ & -1.123 & 0.635$^{***}$ \\ 
  & & (0.120) & (0.170) & (0.226) & (0.182) \\ 
  & & & & & \\ 
 \hline \\[-1.8ex] \\
 Observations & & \multicolumn{1}{c}{104} & \multicolumn{1}{c}{20} & \multicolumn{1}{c}{12} & \multicolumn{1}{c}{12}  \\ 
 R$^{2}$ & & \multicolumn{1}{c}{0.702} & \multicolumn{1}{c}{0.809} & \multicolumn{1}{c}{0.872} & \multicolumn{1}{c}{0.426}  \\ 
 Adjusted R$^{2}$ & & \multicolumn{1}{c}{0.699} & \multicolumn{1}{c}{0.798} & \multicolumn{1}{c}{0.859} & \multicolumn{1}{c}{0.369} \\ 
 Residual Std. Error & & \multicolumn{1}{c}{0.062 (df = 102)} & \multicolumn{1}{c}{0.031 (df = 18)} & \multicolumn{1}{c}{0.080 (df = 10)} & \multicolumn{1}{c}{0.058 (df = 10)} \\ 
 F Statistic & & \multicolumn{1}{c}{239.854$^{***}$ (df = 1; 102)} & \multicolumn{1}{c}{76.016$^{***}$ (df = 1; 18)} & \multicolumn{1}{c}{68.014$^{***}$ (df = 1; 10)} & \multicolumn{1}{c}{7.421$^{**}$ (df = 1; 10)}  \\ \\
 \hline \\
 Implied\ Polynomial\ Scaling\ Factor & & 10.8 & 16.0 & 10.5 & 37.2 \\
 95\%\ Confidence\ Interval & & 9.5\ $-$\ 12.3 & 13.0\ $-$\ 21.3 & 8.3\ $-$\ 14.3 & 20.4\ $-$\ 200 \\ \\
\hline \\
\multicolumn{6}{r}{$^{*}$p$<$0.1; $^{**}$p$<$0.05; $^{***}$p$<$0.01}
\\
 \multicolumn{6}{r}{\textit{Note: Network Operations is normalized relative to the $2012$ AlexNet model.}}
 \end{tabular}
}
\label{tab:deep_learning_regressions}
\end{table*}

It is known that there have been substantial improvements in the efficiency of deep learning training \cite{openai2020}.  In specification ($2$) we introduce a time trend to proxy for these algorithmic changes and find that it increases the explanatory power of the model by $11\%$.  As in previous work, we find clear evidence of efficiency gains: $3$ years of algorithm improvement is equivalent to an increase in computing power of $10\times$.  But, even after accounting for algorithm improvement, we continue to observe a power law between computing power that performance.  This implies that every year deep learning system designers are both taking advantage of year-over-year algorithm improvement and also scaling their models according to the performance trade-offs that we have identified.  And thus, while it is encouraging that algorithmic efficiency has improved, it does not alleviate the inflation in computational burden that we observe.\footnote{In \ref{sec:lessening_burden} we revisit this issue of efficiency improvement as part of a larger discussion about the economic and environmental implications of this rapid rise in the computation needed.}

In specification ($3$) we test a functional form where computation scales exponentially with performance, rather than polynomially.  That form also results in a highly statistically significant reliance on computing power, but has less explanatory power, so we retain specification $1$ as our preferred form.

In specification ($4$) we analyze whether focusing on the best-performing models would yield a different result that looking at all models.  Using a quantile regression, we estimate the scaling at the threshold of the $10\%$ most efficient models.  This estimation shows a similar dependence of performance on computation for these cutting-edge models: \(\mathcal{O}(Performance^{11.9})\).

Thus, in Imagenet, where we have the most data, our baseline analysis and robustness checks paint a strongly coherent picture: deep learning performance improvement is strongly dependent on rapid scale-up of computing power, whether or not algorithmic improvement is accounted for and whether or not one looks at all models or only cutting-edge ones. 

\begin{figure*}[!htb]
 \centering
 \includegraphics[width=.95\textwidth]{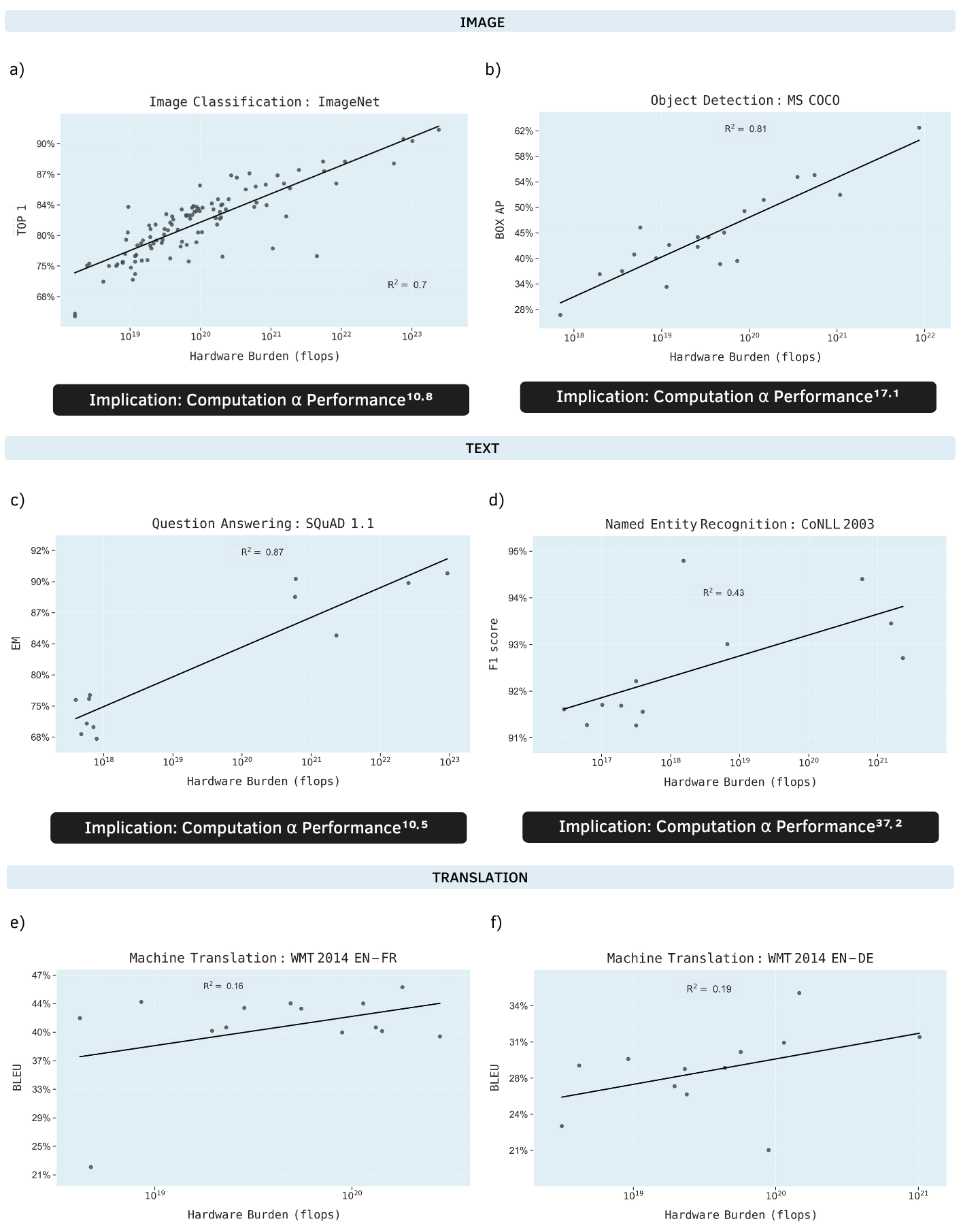}
 \caption{Performance improvement in various deep learning applications as a function of the hardware burden of training that model (in flops).}
 \label{fig:mini_graphs}
\end{figure*}

Despite the efforts of machine learning conferences to encourage more thorough reporting of experimental details (e.g. the reproducibility checklists of ICML \cite{icml} and NeurIPS\cite{sinha2020neurips, pineau2020improving}), few papers in the other benchmark areas provide sufficient information to analyze the computational burden via network operations.  More widely reported, however, are the components needed to calculate an alternative metric: hardware burden.  This also estimates the computation needed, but is less precise since it depends on hardware implementation efficiency.

Table \ref{tab:deep_learning_regressions}b and Figure \ref{fig:mini_graphs} shows progress in the areas of image classification, object detection, question answering, and named entity recognition.  We find highly-statistically significant slopes and strong explanatory power ($R^2$ between $42\%$ and $87\%$) for all benchmarks. Interpreting the coefficients for the five remaining benchmarks shows a slightly lower polynomial dependence for imagenet when calculated using this method ($\approx10.8$), and a dependence of $10.5$ and $17.1$ for question answering and object detection respectively.  Named-entity recognition shows large increases in hardware burden with relatively small improvements in outcomes, implying dependencies of around $37.2$, although this model explains only $43\%$ of the variance so this result should be interpreted as preliminary.  In machine translation we also observe a correlation between compute and performance, but there has not been enough variation in computing power for us to reliably estimate the slope.  

In the supplementary materials, we also test other functional forms for estimating hardware burden. As with the network operations analysis, we find that polynomial models best explain the data, but that exponential models are also plausible.

Collectively, our results make it clear that, across many areas of deep learning, progress in training better models has depended on large increases in the amount of computing power being used.  A dependence on computing power for improved performance is not unique to deep learning, but has also been seen in other areas such as weather prediction, computer chess, computer Go, and oil exploration \cite{thompson2020exponential}.  In those areas there has been enormous growth in the cost of systems, with many cutting-edge models now requiring some of the largest computer systems in the world \cite{bbc2021}.  This could well be deep learning's fate if current trends continue.

\subsection{Future}\label{sec:future}

In this section, we extrapolate the estimates from each domain to understand the projected computational power needed to train models to hit various benchmark performance levels.  To make these targets tangible, we present them not only in terms of the computational power required, but also in terms of the economic and environmental cost of training such models on current hardware.\footnote{Economic and environmental costs are measured using the methodology provided by \cite{strubell2019energy} for V100 GPU training. We confirm pricing as of 2022 based on \cite{google2022gpuprice} and use updated carbon emissions estimates from \cite{dodge2022measuring} (for which we take the geometric mean). These carbon estimates are similar to those provided by \cite{patterson2022carbon, patterson2022footprint}.}
These projections reinforce the growing concern that deep learning's current trajectory will have important negative effects \cite{martinezplumed2018accounting, GARCIAMARTIN201975}.  In this analysis, we focus on the training costs of these models but in section \ref{sec:lessening_burden} we also discuss the deployment cost of these models.  Because the polynomial and exponential functional forms explored in previous section have roughly equivalent statistical fits --- but quite different extrapolations --- we report both in table \ref{tab:projection_analysis}.  

\begin{table*}[!htbp] \centering 
\caption{Implications of achieving performance benchmarks on the computation (in flops), carbon emissions (lbs), and economic costs (\$USD) from deep learning based on projections from polynomial and exponential models.}
\resizebox{\textwidth}{!}{
\begin{tabular}{|c|c|ccc|ccc|}
\hline
{\color[HTML]{000000} } &
  {\color[HTML]{000000} } &
  \multicolumn{3}{c|}{{\color[HTML]{000000} }} &
  \multicolumn{3}{c|}{{\color[HTML]{000000} }} \\
{\color[HTML]{000000} } &
  {\color[HTML]{000000} } &
  \multicolumn{3}{c|}{{\color[HTML]{000000} }} &
  \multicolumn{3}{c|}{{\color[HTML]{000000} }} \\
{\color[HTML]{000000} } &
  {\color[HTML]{000000} } &
  \multicolumn{3}{c|}{\multirow{-3}{*}{{\color[HTML]{000000} \textbf{Polynomial}}}} &
  \multicolumn{3}{c|}{\multirow{-3}{*}{{\color[HTML]{000000} \textbf{Exponential}}}} \\ \cline{3-8} 
{\color[HTML]{000000} } &
  {\color[HTML]{000000} } &
  \multicolumn{1}{c|}{{\color[HTML]{000000} }} &
  \multicolumn{1}{c|}{{\color[HTML]{000000} }} &
  {\color[HTML]{000000} } &
  \multicolumn{1}{c|}{{\color[HTML]{000000} }} &
  \multicolumn{1}{c|}{{\color[HTML]{000000} }} &
  {\color[HTML]{000000} } \\
{\color[HTML]{000000} } &
  {\color[HTML]{000000} } &
  \multicolumn{1}{c|}{{\color[HTML]{000000} }} &
  \multicolumn{1}{c|}{{\color[HTML]{000000} }} &
  {\color[HTML]{000000} } &
  \multicolumn{1}{c|}{{\color[HTML]{000000} }} &
  \multicolumn{1}{c|}{{\color[HTML]{000000} }} &
  {\color[HTML]{000000} } \\
\multirow{-6}{*}{{\color[HTML]{000000} \textbf{Benchmark}}} &
  \multirow{-6}{*}{{\color[HTML]{000000} \textbf{Error Rate}}} &
  \multicolumn{1}{c|}{\multirow{-3}{*}{{\color[HTML]{000000} \textbf{\begin{tabular}[c]{@{}c@{}}Computation\\ Required (flops)\end{tabular}}}}} &
  \multicolumn{1}{c|}{\multirow{-3}{*}{{\color[HTML]{000000} \textbf{\begin{tabular}[c]{@{}c@{}}Environmental\\ Cost (CO$_{2}$)\end{tabular}}}}} &
  \multirow{-3}{*}{{\color[HTML]{000000} \textbf{\begin{tabular}[c]{@{}c@{}}Economic\\ Cost (\$)\end{tabular}}}} &
  \multicolumn{1}{c|}{\multirow{-3}{*}{{\color[HTML]{000000} \textbf{\begin{tabular}[c]{@{}c@{}}Computation\\ Required (flops)\end{tabular}}}}} &
  \multicolumn{1}{c|}{\multirow{-3}{*}{{\color[HTML]{000000} \textbf{\begin{tabular}[c]{@{}c@{}}Environmental\\ Cost (CO$_{2}$)\end{tabular}}}}} &
  \multirow{-3}{*}{{\color[HTML]{000000} \textbf{\begin{tabular}[c]{@{}c@{}}Economic\\ Cost (\$)\end{tabular}}}} \\ \hline \hline
{\color[HTML]{000000} } &
  {\color[HTML]{000000} } &
  \multicolumn{1}{c|}{{\color[HTML]{000000} }} &
  \multicolumn{1}{c|}{{\color[HTML]{000000} }} &
  {\color[HTML]{000000} } &
  \multicolumn{1}{c|}{{\color[HTML]{000000} }} &
  \multicolumn{1}{c|}{{\color[HTML]{000000} }} &
  {\color[HTML]{000000} } \\
{\color[HTML]{000000} } &
  \multirow{-2}{*}{{\color[HTML]{000000} Today: 9.00\%}} &
  \multicolumn{1}{c|}{\multirow{-2}{*}{{\color[HTML]{000000} $10^{23}$}}} &
  \multicolumn{1}{c|}{\multirow{-2}{*}{{\color[HTML]{000000} $10^{5}$}}} &
  \multirow{-2}{*}{{\color[HTML]{000000} $10^{6}$}} &
  \multicolumn{1}{c|}{\multirow{-2}{*}{{\color[HTML]{000000} $10^{24}$}}} &
  \multicolumn{1}{c|}{\multirow{-2}{*}{{\color[HTML]{000000} $10^{6}$}}} &
  \multirow{-2}{*}{{\color[HTML]{000000} $10^{7}$}} \\ \cline{2-8} 
{\color[HTML]{000000} } &
  {\color[HTML]{000000} } &
  \multicolumn{1}{c|}{{\color[HTML]{000000} }} &
  \multicolumn{1}{c|}{{\color[HTML]{000000} }} &
  {\color[HTML]{000000} } &
  \multicolumn{1}{c|}{{\color[HTML]{000000} }} &
  \multicolumn{1}{c|}{{\color[HTML]{000000} }} &
  {\color[HTML]{000000} } \\
{\color[HTML]{000000} } &
  \multirow{-2}{*}{{\color[HTML]{000000} Target 1: 5\%}} &
  \multicolumn{1}{c|}{\multirow{-2}{*}{{\color[HTML]{000000} $10^{26}$}}} &
  \multicolumn{1}{c|}{\multirow{-2}{*}{{\color[HTML]{000000} $10^{8}$}}} &
  \multirow{-2}{*}{{\color[HTML]{000000} $10^{9}$}} &
  \multicolumn{1}{c|}{\multirow{-2}{*}{{\color[HTML]{000000} $10^{30}$}}} &
  \multicolumn{1}{c|}{\multirow{-2}{*}{{\color[HTML]{000000} $10^{13}$}}} &
  \multirow{-2}{*}{{\color[HTML]{000000} $10^{14}$}} \\ \cline{2-8} 
{\color[HTML]{000000} } &
  {\color[HTML]{000000} } &
  \multicolumn{1}{c|}{{\color[HTML]{000000} }} &
  \multicolumn{1}{c|}{{\color[HTML]{000000} }} &
  {\color[HTML]{000000} } &
  \multicolumn{1}{c|}{{\color[HTML]{000000} }} &
  \multicolumn{1}{c|}{{\color[HTML]{000000} }} &
  {\color[HTML]{000000} } \\
\multirow{-6}{*}{{\color[HTML]{000000} ImageNet}} &
  \multirow{-2}{*}{{\color[HTML]{000000} Target 2: 1\%}} &
  \multicolumn{1}{c|}{\multirow{-2}{*}{{\color[HTML]{000000} $10^{33}$}}} &
  \multicolumn{1}{c|}{\multirow{-2}{*}{{\color[HTML]{000000} $10^{16}$}}} &
  \multirow{-2}{*}{{\color[HTML]{000000} $10^{16}$}} &
  \multicolumn{1}{c|}{\multirow{-2}{*}{{\color[HTML]{000000} $10^{92}$}}} &
  \multicolumn{1}{c|}{\multirow{-2}{*}{{\color[HTML]{000000} $10^{74}$}}} &
  \multirow{-2}{*}{{\color[HTML]{000000} $10^{75}$}} \\ \hline
{\color[HTML]{000000} } &
  {\color[HTML]{000000} } &
  \multicolumn{1}{c|}{{\color[HTML]{000000} }} &
  \multicolumn{1}{c|}{{\color[HTML]{000000} }} &
  {\color[HTML]{000000} } &
  \multicolumn{1}{c|}{{\color[HTML]{000000} }} &
  \multicolumn{1}{c|}{{\color[HTML]{000000} }} &
  {\color[HTML]{000000} } \\
{\color[HTML]{000000} } &
  \multirow{-2}{*}{{\color[HTML]{000000} Today: 38.7\%}} &
  \multicolumn{1}{c|}{\multirow{-2}{*}{{\color[HTML]{000000} $10^{22}$}}} &
  \multicolumn{1}{c|}{\multirow{-2}{*}{{\color[HTML]{000000} $10^{4}$}}} &
  \multirow{-2}{*}{{\color[HTML]{000000} $10^{5}$}} &
  \multicolumn{1}{c|}{\multirow{-2}{*}{{\color[HTML]{000000} $10^{22}$}}} &
  \multicolumn{1}{c|}{\multirow{-2}{*}{{\color[HTML]{000000} $10^{4}$}}} &
  \multirow{-2}{*}{{\color[HTML]{000000} $10^{5}$}} \\ \cline{2-8} 
{\color[HTML]{000000} } &
  {\color[HTML]{000000} } &
  \multicolumn{1}{c|}{{\color[HTML]{000000} }} &
  \multicolumn{1}{c|}{{\color[HTML]{000000} }} &
  {\color[HTML]{000000} } &
  \multicolumn{1}{c|}{{\color[HTML]{000000} }} &
  \multicolumn{1}{c|}{{\color[HTML]{000000} }} &
  {\color[HTML]{000000} } \\
{\color[HTML]{000000} } &
  \multirow{-2}{*}{{\color[HTML]{000000} Target 1: 30\%}} &
  \multicolumn{1}{c|}{\multirow{-2}{*}{{\color[HTML]{000000} $10^{23}$}}} &
  \multicolumn{1}{c|}{\multirow{-2}{*}{{\color[HTML]{000000} $10^{6}$}}} &
  \multirow{-2}{*}{{\color[HTML]{000000} $10^{6}$}} &
  \multicolumn{1}{c|}{\multirow{-2}{*}{{\color[HTML]{000000} $10^{24}$}}} &
  \multicolumn{1}{c|}{\multirow{-2}{*}{{\color[HTML]{000000} $10^{7}$}}} &
  \multirow{-2}{*}{{\color[HTML]{000000} $10^{8}$}} \\ \cline{2-8} 
{\color[HTML]{000000} } &
  {\color[HTML]{000000} } &
  \multicolumn{1}{c|}{{\color[HTML]{000000} }} &
  \multicolumn{1}{c|}{{\color[HTML]{000000} }} &
  {\color[HTML]{000000} } &
  \multicolumn{1}{c|}{{\color[HTML]{000000} }} &
  \multicolumn{1}{c|}{{\color[HTML]{000000} }} &
  {\color[HTML]{000000} } \\
\multirow{-6}{*}{{\color[HTML]{000000} MS COCO}} &
  \multirow{-2}{*}{{\color[HTML]{000000} Target 2: 10\%}} &
  \multicolumn{1}{c|}{\multirow{-2}{*}{{\color[HTML]{000000} $10^{31}$}}} &
  \multicolumn{1}{c|}{\multirow{-2}{*}{{\color[HTML]{000000} $10^{13}$}}} &
  \multirow{-2}{*}{{\color[HTML]{000000} $10^{14}$}} &
  \multicolumn{1}{c|}{\multirow{-2}{*}{{\color[HTML]{000000} $10^{49}$}}} &
  \multicolumn{1}{c|}{\multirow{-2}{*}{{\color[HTML]{000000} $10^{31}$}}} &
  \multirow{-2}{*}{{\color[HTML]{000000} $10^{32}$}} \\ \hline
{\color[HTML]{000000} } &
  {\color[HTML]{000000} } &
  \multicolumn{1}{c|}{{\color[HTML]{000000} }} &
  \multicolumn{1}{c|}{{\color[HTML]{000000} }} &
  {\color[HTML]{000000} } &
  \multicolumn{1}{c|}{{\color[HTML]{000000} }} &
  \multicolumn{1}{c|}{{\color[HTML]{000000} }} &
  {\color[HTML]{000000} } \\
{\color[HTML]{000000} } &
  \multirow{-2}{*}{{\color[HTML]{000000} Today: 9.4\%}} &
  \multicolumn{1}{c|}{\multirow{-2}{*}{{\color[HTML]{000000} $10^{22}$}}} &
  \multicolumn{1}{c|}{\multirow{-2}{*}{{\color[HTML]{000000} $10^{4}$}}} &
  \multirow{-2}{*}{{\color[HTML]{000000} $10^{5}$}} &
  \multicolumn{1}{c|}{\multirow{-2}{*}{{\color[HTML]{000000} $10^{22}$}}} &
  \multicolumn{1}{c|}{\multirow{-2}{*}{{\color[HTML]{000000} $10^{5}$}}} &
  \multirow{-2}{*}{{\color[HTML]{000000} $10^{5}$}} \\ \cline{2-8} 
{\color[HTML]{000000} } &
  {\color[HTML]{000000} } &
  \multicolumn{1}{c|}{{\color[HTML]{000000} }} &
  \multicolumn{1}{c|}{{\color[HTML]{000000} }} &
  {\color[HTML]{000000} } &
  \multicolumn{1}{c|}{{\color[HTML]{000000} }} &
  \multicolumn{1}{c|}{{\color[HTML]{000000} }} &
  {\color[HTML]{000000} } \\
{\color[HTML]{000000} } &
  \multirow{-2}{*}{{\color[HTML]{000000} Target 1: 2\%}} &
  \multicolumn{1}{c|}{\multirow{-2}{*}{{\color[HTML]{000000} $10^{29}$}}} &
  \multicolumn{1}{c|}{\multirow{-2}{*}{{\color[HTML]{000000} $10^{11}$}}} &
  \multirow{-2}{*}{{\color[HTML]{000000} $10^{12}$}} &
  \multicolumn{1}{c|}{\multirow{-2}{*}{{\color[HTML]{000000} $10^{51}$}}} &
  \multicolumn{1}{c|}{\multirow{-2}{*}{{\color[HTML]{000000} $10^{34}$}}} &
  \multirow{-2}{*}{{\color[HTML]{000000} $10^{34}$}} \\ \cline{2-8} 
{\color[HTML]{000000} } &
  {\color[HTML]{000000} } &
  \multicolumn{1}{c|}{{\color[HTML]{000000} }} &
  \multicolumn{1}{c|}{{\color[HTML]{000000} }} &
  {\color[HTML]{000000} } &
  \multicolumn{1}{c|}{{\color[HTML]{000000} }} &
  \multicolumn{1}{c|}{{\color[HTML]{000000} }} &
  {\color[HTML]{000000} } \\
\multirow{-6}{*}{{\color[HTML]{000000} SQuAD 1.1}} &
  \multirow{-2}{*}{{\color[HTML]{000000} Target 2: 1\%}} &
  \multicolumn{1}{c|}{\multirow{-2}{*}{{\color[HTML]{000000} $10^{32}$}}} &
  \multicolumn{1}{c|}{\multirow{-2}{*}{{\color[HTML]{000000} $10^{15}$}}} &
  \multirow{-2}{*}{{\color[HTML]{000000} $10^{15}$}} &
  \multicolumn{1}{c|}{\multirow{-2}{*}{{\color[HTML]{000000} $10^{88}$}}} &
  \multicolumn{1}{c|}{\multirow{-2}{*}{{\color[HTML]{000000} $10^{70}$}}} &
  \multirow{-2}{*}{{\color[HTML]{000000} $10^{71}$}} \\ \hline
{\color[HTML]{000000} } &
  {\color[HTML]{000000} } &
  \multicolumn{1}{c|}{{\color[HTML]{000000} }} &
  \multicolumn{1}{c|}{{\color[HTML]{000000} }} &
  {\color[HTML]{000000} } &
  \multicolumn{1}{c|}{{\color[HTML]{000000} }} &
  \multicolumn{1}{c|}{{\color[HTML]{000000} }} &
  {\color[HTML]{000000} } \\
{\color[HTML]{000000} } &
  \multirow{-2}{*}{{\color[HTML]{000000} Today: 5.4\%}} &
  \multicolumn{1}{c|}{\multirow{-2}{*}{{\color[HTML]{000000} $10^{23}$}}} &
  \multicolumn{1}{c|}{\multirow{-2}{*}{{\color[HTML]{000000} $10^{5}$}}} &
  \multirow{-2}{*}{{\color[HTML]{000000} $10^{6}$}} &
  \multicolumn{1}{c|}{\multirow{-2}{*}{{\color[HTML]{000000} $10^{24}$}}} &
  \multicolumn{1}{c|}{\multirow{-2}{*}{{\color[HTML]{000000} $10^{6}$}}} &
  \multirow{-2}{*}{{\color[HTML]{000000} $10^{7}$}} \\ \cline{2-8} 
{\color[HTML]{000000} } &
  {\color[HTML]{000000} } &
  \multicolumn{1}{c|}{{\color[HTML]{000000} }} &
  \multicolumn{1}{c|}{{\color[HTML]{000000} }} &
  {\color[HTML]{000000} } &
  \multicolumn{1}{c|}{{\color[HTML]{000000} }} &
  \multicolumn{1}{c|}{{\color[HTML]{000000} }} &
  {\color[HTML]{000000} } \\
{\color[HTML]{000000} } &
  \multirow{-2}{*}{{\color[HTML]{000000} Target 1: 2\%}} &
  \multicolumn{1}{c|}{\multirow{-2}{*}{{\color[HTML]{000000} $10^{39}$}}} &
  \multicolumn{1}{c|}{\multirow{-2}{*}{{\color[HTML]{000000} $10^{22}$}}} &
  \multirow{-2}{*}{{\color[HTML]{000000} $10^{22}$}} &
  \multicolumn{1}{c|}{\multirow{-2}{*}{{\color[HTML]{000000} $10^{61}$}}} &
  \multicolumn{1}{c|}{\multirow{-2}{*}{{\color[HTML]{000000} $10^{43}$}}} &
  \multirow{-2}{*}{{\color[HTML]{000000} $10^{44}$}} \\ \cline{2-8} 
{\color[HTML]{000000} } &
  {\color[HTML]{000000} } &
  \multicolumn{1}{c|}{{\color[HTML]{000000} }} &
  \multicolumn{1}{c|}{{\color[HTML]{000000} }} &
  {\color[HTML]{000000} } &
  \multicolumn{1}{c|}{{\color[HTML]{000000} }} &
  \multicolumn{1}{c|}{{\color[HTML]{000000} }} &
  {\color[HTML]{000000} } \\
\multirow{-6}{*}{{\color[HTML]{000000} CoNLL 2003}} &
  \multirow{-2}{*}{{\color[HTML]{000000} Target 2: 1\%}} &
  \multicolumn{1}{c|}{\multirow{-2}{*}{{\color[HTML]{000000} $10^{50}$}}} &
  \multicolumn{1}{c|}{\multirow{-2}{*}{{\color[HTML]{000000} $10^{33}$}}} &
  \multirow{-2}{*}{{\color[HTML]{000000} $10^{33}$}} &
  \multicolumn{1}{c|}{\multirow{-2}{*}{{\color[HTML]{000000} $10^{120}$}}} &
  \multicolumn{1}{c|}{\multirow{-2}{*}{{\color[HTML]{000000} $10^{102}$}}} &
  \multirow{-2}{*}{{\color[HTML]{000000} $10^{103}$}} \\ \hline
\end{tabular}}
\label{tab:projection_analysis}
\end{table*}

\textbf{We do not anticipate that the computational requirements implied by the targets in Figure \ref{tab:projection_analysis} will be hit.}  The hardware, environmental, and monetary costs would be prohibitive and enormous effort is going into finding ways to improving scaling performance to avoid these outcomes (as we discuss in the next section).  Nevertheless, these projections do provide a scale for the efficiency improvements that would be needed to hit these performance targets.  For example, even in the more-optimistic model, an additional $567\times$ more computing would be needed to get to an error rate of $5\%$ for ImageNet.  Hitting this in an economical way will require more efficient hardware, more efficient algorithms, or other improvements such that the net impact is at least this large.  Figure \ref{fig:scaling_graphs} (a) shows the large effects that improving the polynomial scaling performance would have on projections and how well these agree with current data.

The rapid escalation in computing needs in Table \ref{tab:projection_analysis} also makes a stronger statement: without substantial efficiency improvements, it will not be possible for deep learning to hit these benchmarks.  Instead, fundamental rearchitecting is needed to lower the computational intensity so that the scaling of these problems becomes less onerous.  We discuss this in detail in section \ref{sec:lessening_burden}, but first we consider how our findings differ from other scaling studies that have been done on deep learning.

\section{Comparison to other scaling studies}

The key question for the future of deep learning is how performance scales up, that is, how much performance \textit{for the field} improves as computing power \textit{increases}. This article addresses this question differently from the growing body of work on deep learning scaling.  In studying the performance of the whole field, rather than just a class of models, we are tracking not just mathematical scaling of models but also the pace of innovation as researchers find better ways to harness computing power, which Rich Sutton has argued is ``the only thing that matters in the long run.'' \cite{Sutton2019}.  By measuring the average progress of the field (or in Table \ref{tab:deep_learning_regressions} specification 4, just the state-of-the-art) we capture cross-researcher, cross-model differences that other scaling papers miss.  Not surprisingly, this results in different estimates for scaling performance.

\begin{table*}[t]
\centering
\caption{Comparison of articles analyzing deep learning scaling.}
\label{tab:lit_review}
\resizebox{\textwidth}{!}{%
\begin{tabular}{|l|cc|ccc|cc|ccc|}
\hline
\rowcolor[HTML]{EFEFEF} 
\cellcolor[HTML]{EFEFEF} &
  \multicolumn{2}{c|}{\cellcolor[HTML]{EFEFEF}} &
  \multicolumn{3}{c|}{\cellcolor[HTML]{EFEFEF}} &
  \multicolumn{2}{c|}{\cellcolor[HTML]{EFEFEF}} &
  \multicolumn{3}{c|}{\cellcolor[HTML]{EFEFEF}} \\
\rowcolor[HTML]{EFEFEF} 
\cellcolor[HTML]{EFEFEF} &
  \multicolumn{2}{c|}{\multirow{-2}{*}{\cellcolor[HTML]{EFEFEF}\textbf{Calculation}}} &
  \multicolumn{3}{c|}{\multirow{-2}{*}{\cellcolor[HTML]{EFEFEF}\textbf{Specificity of Comparison}}} &
  \multicolumn{2}{c|}{\multirow{-2}{*}{\cellcolor[HTML]{EFEFEF}\textbf{Scaling}}} &
  \multicolumn{3}{c|}{\multirow{-2}{*}{\cellcolor[HTML]{EFEFEF}\textbf{Trends}}} \\ \cline{2-11} 
\rowcolor[HTML]{EFEFEF} 
\cellcolor[HTML]{EFEFEF} &
  \multicolumn{1}{c|}{\cellcolor[HTML]{EFEFEF}} &
  \cellcolor[HTML]{EFEFEF} &
  \multicolumn{1}{c|}{\cellcolor[HTML]{EFEFEF}} &
  \multicolumn{1}{c|}{\cellcolor[HTML]{EFEFEF}} &
  \cellcolor[HTML]{EFEFEF} &
  \multicolumn{1}{c|}{\cellcolor[HTML]{EFEFEF}} &
  \cellcolor[HTML]{EFEFEF} &
  \multicolumn{1}{c|}{\cellcolor[HTML]{EFEFEF}} &
  \multicolumn{1}{c|}{\cellcolor[HTML]{EFEFEF}} &
  \cellcolor[HTML]{EFEFEF} \\ [1em]
\rowcolor[HTML]{EFEFEF} 
\multirow{-4}{*}{\cellcolor[HTML]{EFEFEF}\textbf{Paper}} &
  \multicolumn{1}{c|}{\multirow{-2}{*}{\cellcolor[HTML]{EFEFEF}\textbf{Training}}} &
  \multirow{-2}{*}{\cellcolor[HTML]{EFEFEF}\textbf{Inference}} &
  \multicolumn{1}{c|}{\multirow{-2}{*}{\cellcolor[HTML]{EFEFEF}\textbf{\begin{tabular}{c} Across\\ Domains\end{tabular}}}} &
  \multicolumn{1}{c|}{\multirow{-2}{*}{\cellcolor[HTML]{EFEFEF}\textbf{\begin{tabular}{c} Single Domain \\ (e.g. NLP)\end{tabular}}}} &
  \multirow{-2}{*}{\cellcolor[HTML]{EFEFEF}\textbf{\begin{tabular}{c} Single Benchmark \\ (e.g. ImageNet)\end{tabular}}} &
  \multicolumn{1}{c|}{\multirow{-2}{*}{\cellcolor[HTML]{EFEFEF}\textbf{\begin{tabular}{c} Within\\ Model\end{tabular}}}} &
  \multirow{-2}{*}{\cellcolor[HTML]{EFEFEF}\textbf{\begin{tabular}{c} Across\\ Models\end{tabular}}} &
  \multicolumn{1}{c|}{\multirow{-2}{*}{\cellcolor[HTML]{EFEFEF}\textbf{Performance}}} &
  \multicolumn{1}{c|}{\multirow{-2}{*}{\cellcolor[HTML]{EFEFEF}\textbf{Compute}}} &
  \multirow{-2}{*}{\cellcolor[HTML]{EFEFEF}\textbf{\begin{tabular}{c} Performance \\ vs. Compute\end{tabular}}} \\ [1em] \hline 
    \rowcolor[HTML]{6c757d}
    & \multicolumn{1}{c|}{} & & \multicolumn{1}{c|}{} & \multicolumn{1}{c|}{} & & \multicolumn{1}{c|}{} & & \multicolumn{1}{c|}{} & \multicolumn{1}{c|}{} &
   \\
\rowcolor[HTML]{6c757d} 
{\color[HTML]{FFFFFF} Ours (2020 - updated in 2022)} &
  \multicolumn{1}{c|}{\cellcolor[HTML]{6c757d}{\color[HTML]{FFFFFF} \large \ding{56}}} &
  {\color[HTML]{FFFFFF} } &
  \multicolumn{1}{c|}{\cellcolor[HTML]{6c757d}{\color[HTML]{FFFFFF} \large \ding{56} }} &
  \multicolumn{1}{c|}{\cellcolor[HTML]{6c757d}{\color[HTML]{FFFFFF} }} &
  {\color[HTML]{FFFFFF} \large \ding{56}} &
  \multicolumn{1}{c|}{\cellcolor[HTML]{6c757d}{\color[HTML]{FFFFFF} }} &
  {\color[HTML]{FFFFFF} \large \ding{56}} &
  \multicolumn{1}{c|}{\cellcolor[HTML]{6c757d}{\color[HTML]{FFFFFF} \large \ding{56} }} &
  \multicolumn{1}{c|}{\cellcolor[HTML]{6c757d}{\color[HTML]{FFFFFF} \large \ding{56}}} &
  {\color[HTML]{FFFFFF} \large \ding{56}} \\ 
    \rowcolor[HTML]{6c757d}
    & \multicolumn{1}{c|}{} & & \multicolumn{1}{c|}{} & \multicolumn{1}{c|}{} & & \multicolumn{1}{c|}{} & & \multicolumn{1}{c|}{} & \multicolumn{1}{c|}{} &
   \\ \hline
      & \multicolumn{1}{c|}{} & & \multicolumn{1}{c|}{} & \multicolumn{1}{c|}{} & & \multicolumn{1}{c|}{} & & \multicolumn{1}{c|}{} & \multicolumn{1}{c|}{} &
   \\
AI and Compute (2018) \cite{openai2018} &
  \multicolumn{1}{c|}{\large \ding{56}} &
  \multicolumn{1}{c|}{} &
  \multicolumn{1}{c|}{\large \ding{56}} &
  \multicolumn{1}{c|}{\large \ding{56}} &
   &
  \multicolumn{1}{c|}{} &
  \large \ding{56} &
  \multicolumn{1}{c|}{} &
  \multicolumn{1}{c|}{\large \ding{56}} &
   \\      
   & \multicolumn{1}{c|}{} & & \multicolumn{1}{c|}{} & \multicolumn{1}{c|}{} & & \multicolumn{1}{c|}{} & & \multicolumn{1}{c|}{} & \multicolumn{1}{c|}{} &
   \\ \hline 
     & \multicolumn{1}{c|}{} & & \multicolumn{1}{c|}{} & \multicolumn{1}{c|}{} & & \multicolumn{1}{c|}{} & & \multicolumn{1}{c|}{} & \multicolumn{1}{c|}{} &
   \\
Compute Trends Across Three Eras of Machine Learning (2022) \cite{sevilla2022compute} &
  \multicolumn{1}{c|}{\large \ding{56}} &
  \multicolumn{1}{c|}{} &
  \multicolumn{1}{c|}{\large \ding{56}} &
  \multicolumn{1}{c|}{} &
   &
  \multicolumn{1}{c|}{} &
  \large \ding{56} &
  \multicolumn{1}{c|}{} &
  \multicolumn{1}{c|}{\large \ding{56}} &
   \\      
   & \multicolumn{1}{c|}{} & & \multicolumn{1}{c|}{} & \multicolumn{1}{c|}{} & & \multicolumn{1}{c|}{} & & \multicolumn{1}{c|}{} & \multicolumn{1}{c|}{} &
   \\ \hline
    & \multicolumn{1}{c|}{} & & \multicolumn{1}{c|}{} & \multicolumn{1}{c|}{} & & \multicolumn{1}{c|}{} & & \multicolumn{1}{c|}{} & \multicolumn{1}{c|}{} &
   \\
``AI and Compute'' Trend isn't Predictive of What is Happening (2021) \cite{lyzhov2021ai} &
  \multicolumn{1}{c|}{\large \ding{56}} &
  \multicolumn{1}{c|}{} &
  \multicolumn{1}{c|}{\large \ding{56}} &
  \multicolumn{1}{c|}{} &
   &
  \multicolumn{1}{c|}{} &
  \large \ding{56} &
  \multicolumn{1}{c|}{} &
  \multicolumn{1}{c|}{\large \ding{56}} &
   \\       
    & \multicolumn{1}{c|}{} & & \multicolumn{1}{c|}{} & \multicolumn{1}{c|}{} & & \multicolumn{1}{c|}{} & & \multicolumn{1}{c|}{} & \multicolumn{1}{c|}{} &
   \\ \hline
     & \multicolumn{1}{c|}{} & & \multicolumn{1}{c|}{} & \multicolumn{1}{c|}{} & & \multicolumn{1}{c|}{} & & \multicolumn{1}{c|}{} & \multicolumn{1}{c|}{} &
   \\
\begin{tabular}[c]{@{}l@{}}AI and Compute: How Much Longer Can Computing Power Drive\\  Artificial Intelligence Progress? (2021) \cite{Lohn2022ai} \end{tabular} &
  \multicolumn{1}{c|}{\large \ding{56}} &
  \multicolumn{1}{c|}{} &
  \multicolumn{1}{c|}{\large \ding{56}} &
  \multicolumn{1}{c|}{} &
   &
  \multicolumn{1}{c|}{} &
  \large \ding{56} &
  \multicolumn{1}{c|}{} &
  \multicolumn{1}{c|}{\large \ding{56}} &
   \\ 
      & \multicolumn{1}{c|}{} & & \multicolumn{1}{c|}{} & \multicolumn{1}{c|}{} & & \multicolumn{1}{c|}{} & & \multicolumn{1}{c|}{} & \multicolumn{1}{c|}{} &
   \\
   \hline
      & \multicolumn{1}{c|}{} & & \multicolumn{1}{c|}{} & \multicolumn{1}{c|}{} & & \multicolumn{1}{c|}{} & & \multicolumn{1}{c|}{} & \multicolumn{1}{c|}{} &
   \\
Scaling Laws for Neural Language Models (2020) \cite{kaplan2020scaling} &
  \multicolumn{1}{c|}{\large \ding{56}} &
  \multicolumn{1}{c|}{} &
  \multicolumn{1}{c|}{} &
  \multicolumn{1}{c|}{\large \ding{56}} &
   &
  \multicolumn{1}{c|}{} &
  \multicolumn{1}{c|}{\large \ding{56}} &
  \multicolumn{1}{c|}{} &
  \multicolumn{1}{c|}{} & \multicolumn{1}{c|}{\large \ding{56}}
   \\ 
    & \multicolumn{1}{c|}{} & & \multicolumn{1}{c|}{} & \multicolumn{1}{c|}{} & & \multicolumn{1}{c|}{} & & \multicolumn{1}{c|}{} & \multicolumn{1}{c|}{} &
   \\
   \hline
   
      & \multicolumn{1}{c|}{} & & \multicolumn{1}{c|}{} & \multicolumn{1}{c|}{} & & \multicolumn{1}{c|}{} & & \multicolumn{1}{c|}{} & \multicolumn{1}{c|}{} &
   \\
Deep Residual Learning for Image Recognition (2015) \cite{he2015resnet} &
  \multicolumn{1}{c|}{\large \ding{56}} & \multicolumn{1}{c|}{}
   &
  \multicolumn{1}{c|}{} &
  \multicolumn{1}{c|}{} &  \multicolumn{1}{c|}{\large \ding{56}}
   &
  \multicolumn{1}{c|}{\large \ding{56}} &
   &
  \multicolumn{1}{c|}{} &
  \multicolumn{1}{c|}{} & \multicolumn{1}{c|}{\large \ding{56}}
   \\ 
     & \multicolumn{1}{c|}{} & & \multicolumn{1}{c|}{} & \multicolumn{1}{c|}{} & & \multicolumn{1}{c|}{} & & \multicolumn{1}{c|}{} & \multicolumn{1}{c|}{} &
   \\
   \hline
   
      & \multicolumn{1}{c|}{} & & \multicolumn{1}{c|}{} & \multicolumn{1}{c|}{} & & \multicolumn{1}{c|}{} & & \multicolumn{1}{c|}{} & \multicolumn{1}{c|}{} &
   \\
Dual Path Networks (2017) \cite{chen2017dpn} &
  \multicolumn{1}{c|}{\large \ding{56}} & \multicolumn{1}{c|}{}
   &
  \multicolumn{1}{c|}{} &
  \multicolumn{1}{c|}{} & \multicolumn{1}{c|}{\large \ding{56}}
   &
  \multicolumn{1}{c|}{\large \ding{56}} &
   &
  \multicolumn{1}{c|}{} &
  \multicolumn{1}{c|}{} & \multicolumn{1}{c|}{\large \ding{56}}
   \\ 
      & \multicolumn{1}{c|}{} & & \multicolumn{1}{c|}{} & \multicolumn{1}{c|}{} & & \multicolumn{1}{c|}{} & & \multicolumn{1}{c|}{} & \multicolumn{1}{c|}{} &
   \\
   \hline
   
     & \multicolumn{1}{c|}{} & & \multicolumn{1}{c|}{} & \multicolumn{1}{c|}{} & & \multicolumn{1}{c|}{} & & \multicolumn{1}{c|}{} & \multicolumn{1}{c|}{} &
   \\
What is the State of Neural Network Pruning? (2020) \cite{MLSYS2020_d2ddea18} &
  \multicolumn{1}{c|}{\large \ding{56}} & \multicolumn{1}{c|}{}
   &
  \multicolumn{1}{c|}{} &
  \multicolumn{1}{c|}{} & \multicolumn{1}{c|}{\large \ding{56}}
   &
  \multicolumn{1}{c|}{\large \ding{56}} &
   &
  \multicolumn{1}{c|}{} &
  \multicolumn{1}{c|}{} & \multicolumn{1}{c|}{\large \ding{56}}
   \\
     & \multicolumn{1}{c|}{} & & \multicolumn{1}{c|}{} & \multicolumn{1}{c|}{} & & \multicolumn{1}{c|}{} & & \multicolumn{1}{c|}{} & \multicolumn{1}{c|}{} &
   \\
   \hline
      & \multicolumn{1}{c|}{} & & \multicolumn{1}{c|}{} & \multicolumn{1}{c|}{} & & \multicolumn{1}{c|}{} & & \multicolumn{1}{c|}{} & \multicolumn{1}{c|}{} &
   \\
EfficientNet: Rethinking Model Scaling for Convolutional Neural Networks (2019) \cite{tan2019efficientnet} &
  \multicolumn{1}{c|}{\large \ding{56}} &
  \multicolumn{1}{c|}{} &
  \multicolumn{1}{c|}{} &
  \multicolumn{1}{c|}{} &
  \multicolumn{1}{c|}{\large \ding{56}} &
  \multicolumn{1}{c|}{\large \ding{56}} & 
   & 
  \multicolumn{1}{c|}{} &
  \multicolumn{1}{c|}{} & \multicolumn{1}{c|}{\large \ding{56}}
   \\ 
     & \multicolumn{1}{c|}{} & & \multicolumn{1}{c|}{} & \multicolumn{1}{c|}{} & & \multicolumn{1}{c|}{} & & \multicolumn{1}{c|}{} & \multicolumn{1}{c|}{} &
   \\ \hline 
   & \multicolumn{1}{c|}{} & & \multicolumn{1}{c|}{} & \multicolumn{1}{c|}{} & & \multicolumn{1}{c|}{} & & \multicolumn{1}{c|}{} & \multicolumn{1}{c|}{} &
   \\
Learning Transferable Architectures for Scalable Image Recognition (2017) \cite{zoph2017nasnet} &
  \multicolumn{1}{c|}{\large \ding{56}} &
  \multicolumn{1}{c|}{} &
  \multicolumn{1}{c|}{} &
  \multicolumn{1}{c|}{} &
  \multicolumn{1}{c|}{\large \ding{56}} &
  \multicolumn{1}{c|}{\large \ding{56}} &
   &
  \multicolumn{1}{c|}{} &
  \multicolumn{1}{c|}{} & \multicolumn{1}{c|}{\large \ding{56}}
   \\ 
      & \multicolumn{1}{c|}{} & & \multicolumn{1}{c|}{} & \multicolumn{1}{c|}{} & & \multicolumn{1}{c|}{} & & \multicolumn{1}{c|}{} & \multicolumn{1}{c|}{} &
   \\ \hline
    & \multicolumn{1}{c|}{} & & \multicolumn{1}{c|}{} & \multicolumn{1}{c|}{} & & \multicolumn{1}{c|}{} & & \multicolumn{1}{c|}{} & \multicolumn{1}{c|}{} &
   \\
Scaling Laws for Deep Learning (2021) \cite{Jonathan2021Scaling} &
  \multicolumn{1}{c|}{\large \ding{56}} &
  \multicolumn{1}{c|}{} &
  \multicolumn{1}{c|}{} &
  \multicolumn{1}{c|}{} &
  \multicolumn{1}{c|}{\large \ding{56}} &
  \multicolumn{1}{c|}{\large \ding{56}} &
  \multicolumn{1}{c|}{} &
  \multicolumn{1}{c|}{} &
  \multicolumn{1}{c|}{} &
  \multicolumn{1}{c|}{\large \ding{56}}
   \\ 
    & \multicolumn{1}{c|}{} & & \multicolumn{1}{c|}{} & \multicolumn{1}{c|}{} & & \multicolumn{1}{c|}{} & & \multicolumn{1}{c|}{} & \multicolumn{1}{c|}{} &
   \\ \hline
    & \multicolumn{1}{c|}{} & & \multicolumn{1}{c|}{} & \multicolumn{1}{c|}{} & & \multicolumn{1}{c|}{} & & \multicolumn{1}{c|}{} & \multicolumn{1}{c|}{} &
   \\
Deep Learning Scaling is Predictable, Empirically (2017) \cite{Joel2017dl} &
  \multicolumn{1}{c|}{\large \ding{56}} &
  \multicolumn{1}{c|}{} &
  \multicolumn{1}{c|}{} &
  \multicolumn{1}{c|}{} &
  \multicolumn{1}{c|}{\large \ding{56}} &
  \multicolumn{1}{c|}{\large \ding{56}} &
  \multicolumn{1}{c|}{} &
  \multicolumn{1}{c|}{} &
  \multicolumn{1}{c|}{} &
  \multicolumn{1}{c|}{\large \ding{56}}
   \\ 
    & \multicolumn{1}{c|}{} & & \multicolumn{1}{c|}{} & \multicolumn{1}{c|}{} & & \multicolumn{1}{c|}{} & & \multicolumn{1}{c|}{} & \multicolumn{1}{c|}{} &
   \\ \hline
    & \multicolumn{1}{c|}{} & & \multicolumn{1}{c|}{} & \multicolumn{1}{c|}{} & & \multicolumn{1}{c|}{} & & \multicolumn{1}{c|}{} & \multicolumn{1}{c|}{} &
   \\
Compute and Energy Consumption Trends in Deep Learning Inference (2021) \cite{desislavov2021compute} &
  \multicolumn{1}{c|}{} &
  \multicolumn{1}{c|}{\large \ding{56}} &
  \multicolumn{1}{c|}{} &
  \multicolumn{1}{c|}{} &
  \multicolumn{1}{c|}{\large \ding{56}} &
  \multicolumn{1}{c|}{} &
  \multicolumn{1}{c|}{\large \ding{56}} &
  \multicolumn{1}{c|}{\large \ding{56}} &
  \multicolumn{1}{c|}{\large \ding{56}} &
  \multicolumn{1}{c|}{\large \ding{56}}
   \\ 
    & \multicolumn{1}{c|}{} & & \multicolumn{1}{c|}{} & \multicolumn{1}{c|}{} & & \multicolumn{1}{c|}{} & & \multicolumn{1}{c|}{} & \multicolumn{1}{c|}{} &
   \\ \hline
\end{tabular}%
}
\end{table*}

Table \ref{tab:lit_review} summarizes how our analysis compares to other papers in this field.  Like most of the papers in this field, we focus on the training costs needed.  Of these, other papers generally fall into two groups: within-model experiments and across-model historical analyses. Within-model experiments have a specific reference point, the benchmark being analyzed. As a result, they provide an excellent view of how deep learning performance scales as more computation is used. But this analysis has a limited scope --- only the model used by the researchers --- and thus these analyses cannot make any claims about innovation happening across the field. In contrast, most across-model historical analyses can capture innovation, but lack the specificity of benchmark comparisons and thus cannot articulate the performance benefits of additional computation. Our analyses sits between these, capturing both the specificity of individual benchmarks (and thus the performance vs compute trade-off), but also the breadth of looking across models that allows us to capture innovation in the field. More specifically, our approach provides the following benefits:

First, our analysis has more specificity in its comparison set because we examine performance within particular deep learning benchmarks rather than across domains. This contrasts, for example, with \cite{openai2018, sevilla2022compute, Lohn2022ai} that aggregate analysis across different deep learning benchmarks and domains, and therefore do not distinguish between increases in computational burden within particular tasks (e.g. image classification on ImageNet) and the application of deep learning to more computationally-intensive sub-fields (e.g. playing Go).  In contrast with those papers, we are able to measure the growth in computational burden that has been needed to get better performance on particular benchmarks. 

Second, our analysis tracks the evolution of performance in each field differently than \textbf{within model} studies where authors examine trade-offs by implementing many different model configurations \cite{kaplan2020scaling, he2015resnet, chen2017dpn, zoph2017nasnet, tan2019efficientnet, MLSYS2020_d2ddea18}. The weakness of these experimental studies is that the `space' of potential models they explore is limited to what the authors implement themselves --- for example \cite{MLSYS2020_d2ddea18} only considers particular network architectures and \cite{kaplan2020scaling} only studies language models that use the Adam optimizer.  Therefore these experimental approaches necessarily miss innovations that those authors did not consider or that take deep learning in new directions. These exclusions hinder the ability of those scaling studies to capture progress over time, whereas our \textbf{across models} approach is able to capture the innovations that are missed by these experimental studies.

Third, as we discuss in more detail below, our analysis is more reliable for estimating future progress, i.e., how things \textbf{scale up}, because analyses that estimate scaling by looking at how performance deteriorates with less computing power, i.e., by \textbf{scaling down}, can yield artificially rosy estimates.

In general, one would expect that our analysis would show faster performance gains from increased computing power than other studies because while both capture the improvements from scaling within models, our also accounts for innovation over time that could further improve on this performance. In practice, however, this is not what we see.  While some studies do show slower model scaling than we show for the field, others show faster scaling.  We hypothesize that this inconsistency arises because of differences in measurement approaches, depending on whether scaling is determined by how performance scales with additional computation versus how it scales with less.  These approaches sound like they might be symmetric, but they are not.

\begin{figure*}
 \centering
 \includegraphics[width=.9\textwidth]{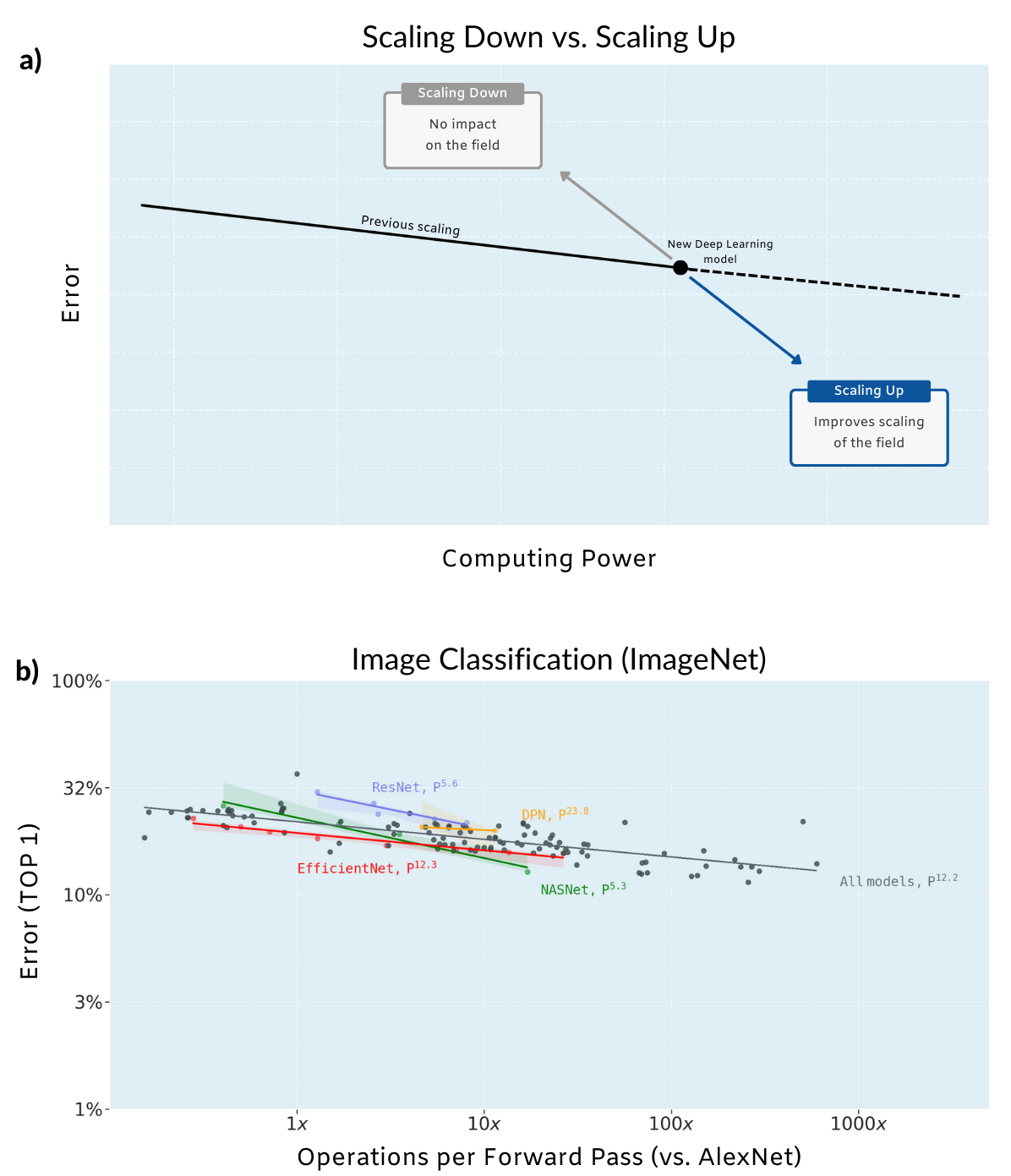}
 \caption{Scaling laws for Deep Learning. \textbf{(a)} schematic representation of scaling up versus scaling down where the slope is the same but the implications are quite different.  \textbf{(b)} Comparison of ImageNet scaling estimates between this paper, ResNet \cite{he2015resnet}, DPN \cite{chen2017dpn}, NASNet \cite{zoph2017nasnet}, and EfficientNet \cite{tan2019efficientnet}.}
 \label{fig:scaling_graphs}
\end{figure*}

Put simply, the `scaling up' approach measures improvements to the state of the art, whereas the `scaling down' approach measures performance deterioration.  To see this, we consider each in turn.  The scaling up approach measures how performance changes as computing power \textit{increases}.  As shown in Figure \ref{fig:scaling_graphs} (a), scaling up clearly measures the progression in the field in an intuitive way: good scaling means that as more computation is used fewer errors are being made and the state of the art is improving.  A scaling up approach naturally emerges from observational analyses, like ours, where both performance and computational power are increasing over time.

Observing this same estimate (i.e. slope) has a less clear implication when scaling down a state-of-the-art model to see how it performs with less computation.  In this latter case, a steep slope (``good scaling'') simply means that smaller models are farther from the frontier of what is possible.  A simple example illustrates how misleading this can be: imagine a state-of-the-art model that, if it used one flop less for training, became useless (i.e. would have an error rate = 100\%).  Such a hypothetical system would have an enormous change in performance from a small change in computing, so it would seem to scale enormously well.  But, in reality, the system simply breaks when it uses less computation.  A less extreme example of this seems to be the case with ResNet scaling\cite{he2015resnet}, which nominally shows dramatically better scaling than the field as a whole (i.e., to get a performance improvement requires compute to only grows to the power of $5.6$, whereas for the field it grows by $12.2$).  In practice, however, these models are not competitive for state-of-the-art performance (as shown in Figure \ref{fig:scaling_graphs}) and the rapid scaling just seems to indicate that their relative performance deteriorates more quickly as less computation is used.  

The example of NasNets provides an alternative way to understand why we hypothesize that models will not scale up. The class of NasNet models reports to have vastly better scaling than the field as a whole.  If one projects this reported scaling, it would indicate that a (hypothetical) NasNet using the same amount of computation as one of the largest vision models (XCiT-L24) should achieve an error rate of 6.85\%, handily beating the actual state of the art by more than 2\%. That we do not see NasNets easily beating state-of-the-art models is strongly suggestive that they do not in fact scale \emph{up} this well.

The problem of misinterpreting rapidly deteriorating performance as an indication of good scaling is not limited to deep learning.  Similar issues arise in studies of parallelism and lead to perverse conclusions including a parallel algorithm that `scales well' being run on 128 cores and being outperformed by a serial algorithm running on a single core \cite{McSherry2015}.  To avoid having deep learning studies fall prey to this same defect, it is important to analyze how models \textit{scale up} and to compare such performance \textit{across} models so that deteriorating performance is not misinterpreted as a virtue. 

With this conceptual framework, we can consider how our results compare to other scaling studies.  Our results notionally agree with the power law growths found by the experimental studies of \cite{kaplan2020scaling} and \cite{hernandez2021scaling}, and the rapid growth (appearing to be approximately exponential) indicated by the plots in \cite{MLSYS2020_d2ddea18}.  But we can also be more quantitative, comparing our ImageNet scaling results to those found by others \cite{he2015resnet, chen2017dpn, zoph2017nasnet, tan2019efficientnet}.  Figure \ref{fig:scaling_graphs} (b) plots these studies, and ours, on the same graph.\footnote{Because the comparison papers do not report all the necessary information to calculate network operations, we instead measure computing by the operations per network pass for this analysis.}

As already mentioned, two sets of analyses show faster scaling for individual models that we observe for the field: ResNet and NASNet.  NASNet in particular achieves cutting-edge performance and shows an impressive scaling of $\mathcal{O}(Performance^{5.3})$.  Based on the argument above, however, we would expect that observing scaling faster than the whole field is just an indication that performance deteriorates rapidly for smaller models.

Of the other scaling studies that we consider, perhaps the most interesting is EfficientNet \cite{tan2019efficientnet}.  For most levels of computation, EfficientNets are at, or close to, the frontier of performance.  So they do not seem to fall prey to the `rapid deterioration equals good scaling' trap.  Moreover, while EfficientNet scales less well than the field as a whole, it is very close ($p^{12.3}$ versus $p^{12.2}$) and remains near the state-of-the-art for each level of computing.  All of which suggest that it may be harnessing many of the important innovations used across the field.  

Thus, because our analysis accounts for innovation and because it is not misled by overly-optimistic scaling down studies, we believe it provides a better long-term view of the evolution of computing power as higher model performance is sought.

\section{Lessening the Computational Burden}\label{sec:lessening_burden}

The economic and environmental burden of hitting the performance benchmarks in Section \ref{sec:future} suggest that Deep Learning is facing an important challenge: either find a way to increase performance without increasing computing power, or have performance stagnate as computational requirements become a constraint. In this section, we briefly survey approaches that are being used to address this challenge.  As with the rest of the paper, we focus on just the training cost of these models, rather than including the deployment cost since the latter depend on usage and diffusion patterns for which data is not available.  Since total costs must necessarily be higher than just training costs, our analysis provides a lower bound on the total computation needed for any given level of performance.

\textbf{Increasing computing power: Hardware accelerators.} For much of the $2010$s, moving to more-efficient hardware platforms (and more of them) was a key source of increased computing power \cite{thompson2021decline}.  For deep learning, these included mostly GPU and TPU implementations, although it has increasingly also included FPGA and other ASICs.  Fundamentally, all of these approaches sacrifice generality of the computing platform for the efficiency of increased specialization.  

In recent years, hardware specialization has provided gains in both compute per dollar and compute per watt, for example TPUs improved by approximately $1.5\times$ in compute per dollar from 2017 to 2019\cite{jouppi2020domain,jouppi2021ten} and $4.9\times$ in compute per watt from 2017 to 2020 \cite{reuther2019survey,reuther2020survey,reuther2021ai,jouppi2020domain,jouppi2021ten}.  As highlighted in \cite{patterson2021carbon}, this can have big implications for reducing the carbon impact of deep learning \textit{if} one is willing to assume that most models will be trained in state-of-the-art cloud facilities located near green energy sources.

Even with significant gains from specialization so far, it is unclear that specialization will be able to continue to provide such gains in the future since it faces diminishing returns \cite{Leiserson20}.  Other hardware frameworks are being explored as alternatives, including analog hardware with in-memory computation \cite{ambrogio2018equivalent,kim2019confined}, neuromorphic computing \cite{davies2019progress}, optical computing \cite{lin2018all}, and quantum computing based approaches \cite{welser2018future}, as well as hybrid approaches \cite{potok2018study}.  Thus far, however, such attempts have yet to disrupt the GPU/TPU and FPGA/ASIC architectures. Of these, quantum computing is the approach with perhaps the most long-term upside, since it might offer a potential for sustained exponential increases in computing power \cite{gambetta2019cramming,cross2019validating}.

\textbf{Reducing computational complexity: Network Compression and Acceleration.} This body of work primarily focuses on taking a trained neural network and sparsifying or otherwise compressing the connections in the network, so that it requires less computation to use it in prediction tasks \cite{cheng2017survey}. This is typically done by using optimization or heuristics such as ``pruning" weights \cite{dong2017more,MLSYS2020_d2ddea18}, quantizing the network \cite{hubara2016binarized}, or using low-rank compression \cite{wen2017coordinating}, yielding a network that retains the performance of the original network but requires fewer floating point operations to evaluate. Not all results that have claimed success in this field have really achieved it \cite{science2020hutson}, but those that have achieved gains have not been large enough to mitigate the overall orders-of-magnitude increases of computation in the field (e.g. the work \cite{chen2018big} reduces computation by a factor of 2, and \cite{wu2020lite} reduces it by a factor of 8 on a specific NLP architecture, both without reducing performance significantly).\footnote{Some works, e.g. \cite{han2015deep} focus more on the reduction in the memory footprint of the model. \cite{han2015deep} achieved 50x compression.} Furthermore, many of these works focus on improving the computational cost of evaluating the deployed network,\footnote{An exception is \cite{frankle2018lottery}, which shows pruning during training.} which is useful, but does not mitigate the training cost, which can itself be prohibitive.

\textbf{Finding high-performing small deep learning architectures: Neural Architecture Search and Meta Learning.} It has become popular to use optimization to find network architectures that are computationally efficient to train while retaining good performance on some class of learning problems, e.g. \cite{pham2018efficient}, \cite{cai2019once} and \cite{finn2017model}.  Designers exploit the fact that many datasets are similar and therefore information from previously trained models can be used (meta learning \cite{pham2018efficient} and transfer learning \cite{long2017deep}). While often quite successful, the downside is that the current overhead of doing meta learning or neural architecture search is itself computationally intense (since it requires training many models on a wide variety of datasets) \cite{pham2018efficient}.  Promisingly, however, the size of this extra overhead cost has been falling \cite{cai2018proxylessnas, cai2019once}.  

An important limitation to meta learning is the scope of the data that the original model was trained on.  For example, for ImageNet, \cite{Barbu2019} showed that image classification performance depends heavily on image biases (e.g. an object is often photographed at a particular angle with a particular pose), and that \textit{without these biases transfer learning performance drops $45\%$}.  Even with novel data sets purposely built to mimic their training data, \cite{recht2019} finds that performance drops $11-14\%$.
Hence, while there seems to be a number of promising research directions for making deep learning computation grow at a more attractive rate, they have yet to achieve the orders-of-magnitude improvements needed to allow deep learning progress to continue scaling. 

Another possible approach to evade the rising computational burden of deep learning would be to move to other, perhaps as yet undiscovered or underappreciated types of machine learning.  As Figure \ref{fig:theory_template}(b) showed, ``expert'' models can be much more computationally-efficient, but their performance plateaus if those experts cannot see all the contributing factors that a flexible model might explore. One example where such techniques are already outperforming deep learning models are those where engineering and physics knowledge can be more-directly applied: the recognition of known objects (e.g. vehicles) \cite{He2019, Tzoumas2019}, and those using biologically-inspired methods, e.g. for learning neural controller architectures \cite{lechner2020}. The recent development of symbolic approaches to machine learning take this a step further, using symbolic methods to efficiently learn and apply ``expert knowledge" in some sense, e.g. \cite{udrescu2020ai} which learns physics laws from data, or approaches \cite{mao2019neuro,asai2020learning,yi2018neural} which apply neuro-symbolic reasoning to scene understanding, reinforcement learning, and natural language processing tasks, building a high-level symbolic representation of the system in order to be able to understand and explore it more effectively with less data.

A recent study which compared neural to non-neural models of text classification found, unsurprisingly, that non-neural approaches outperformed when data was limited, but that neural approaches won out when data was copious (echoing our discussion in the theory section).  For those researchers, moving to non-neural approaches produced a more than $23{\times}$ speedup with a 5\% drop in performance \cite{CUNHA2021102481}.

Finally, exploring combinations of the above approaches to achieve even larger compounded gains may be fruitful in reducing computation \cite{Dally2020}, or reducing environmental damage\cite{patterson2021carbon}. Based on the reported gains of the individual approaches so far, however, we don't believe compounding them would yet be sufficient to dramatically bring down the very severe scaling we've seen in this work.

\section{Conclusion}
The explosion in computing power used for deep learning models has ended the ``AI winter'' and set new benchmarks for computer performance on a wide range of tasks.  However, deep learning's prodigious appetite for computing power limits how far it can improve performance in its current form, particularly in an era when improvements in hardware performance are slowing.  This article shows that the growing computational burden of deep learning will soon be constraining for a range of applications, making the achievement of important benchmark milestones impossible if current trajectories hold.  Finally, we have discussed the likely impact of these computational limits: forcing Deep Learning towards less computationally-intensive methods of improvement, and pushing machine learning towards techniques that are more computationally-efficient than current deep learning approaches.

\section*{Acknowledgments}
The authors would like to acknowledge funding from the MIT Initiative on the Digital Economy and the Korean Government. This research was partially supported by Basic Science Research Program through the National Research Foundation of Korea(NRF) funded by the Ministry of Science, ICT \& Future Planning(2017R1C1B1010094).  This research was sponsored in part by the United States Air Force Research Laboratory and was accomplished under Cooperative Agreement Number FA8750-19-2-1000.  The views and conclusions contained in this document are those of the authors and should not be interpreted as representing the official policies, either expressed or implied, of the United States Air Force or the U.S. Government.  The U.S. Government is authorized to reproduce and distribute reprints for Government purposes notwithstanding any copyright notation herein.

\bibliographystyle{unsrt}  
\bibliography{references}

\newpage
\section*{Supplemental Materials}\label{supplemental}
\pagenumbering{arabic}
\section{Methodology}\label{supp:method}

\subsection{Data collection}
We collect data on the performance and computational requirements of various deep learning models from 
arXiv (\url{arXiv.org}), which is an open-access archive where scholars upload preprints of their scientific papers (once they are approved by moderators). Preprints are  categorized into the following fields: mathematics, physics, astrophysics, computer science, quantitative biology, quantitative finance, statistics, electrical engineering and systems science, and economics. Moderators review submissions and can (re)categorize them or reject them (but this is not a full peer review), as discussed at \url{https://arxiv.org/corr/subjectclasses}. Preprints are accessible and freely distributed worldwide.

To collect preprints of interest from arXiv, we use the search terms of specific tasks and benchmarks (as discussed below). These allow us to gather pdf pre-prints from \url{arXiv.org}. From each paper, we attempt to gather the following pieces of information:

\begin{itemize}
 \item Application area (e.g. image classification);
 \item Benchmark (name, version, \# of observations in the training set);
 \item Paper details (authors, title, year made public);
 \item Model details (name, performance metrics, training time);
  \item Network characteristics (\# parameters, \# of training epochs, \# floating-point operations and \# multiply-adds per forward pass);
 \item Hardware usage (hardware type, hardware performance (GFLOPs), \# processors);
\end{itemize}

\begin{figure}[!htb]
 \centering
 \includegraphics[width=\columnwidth]{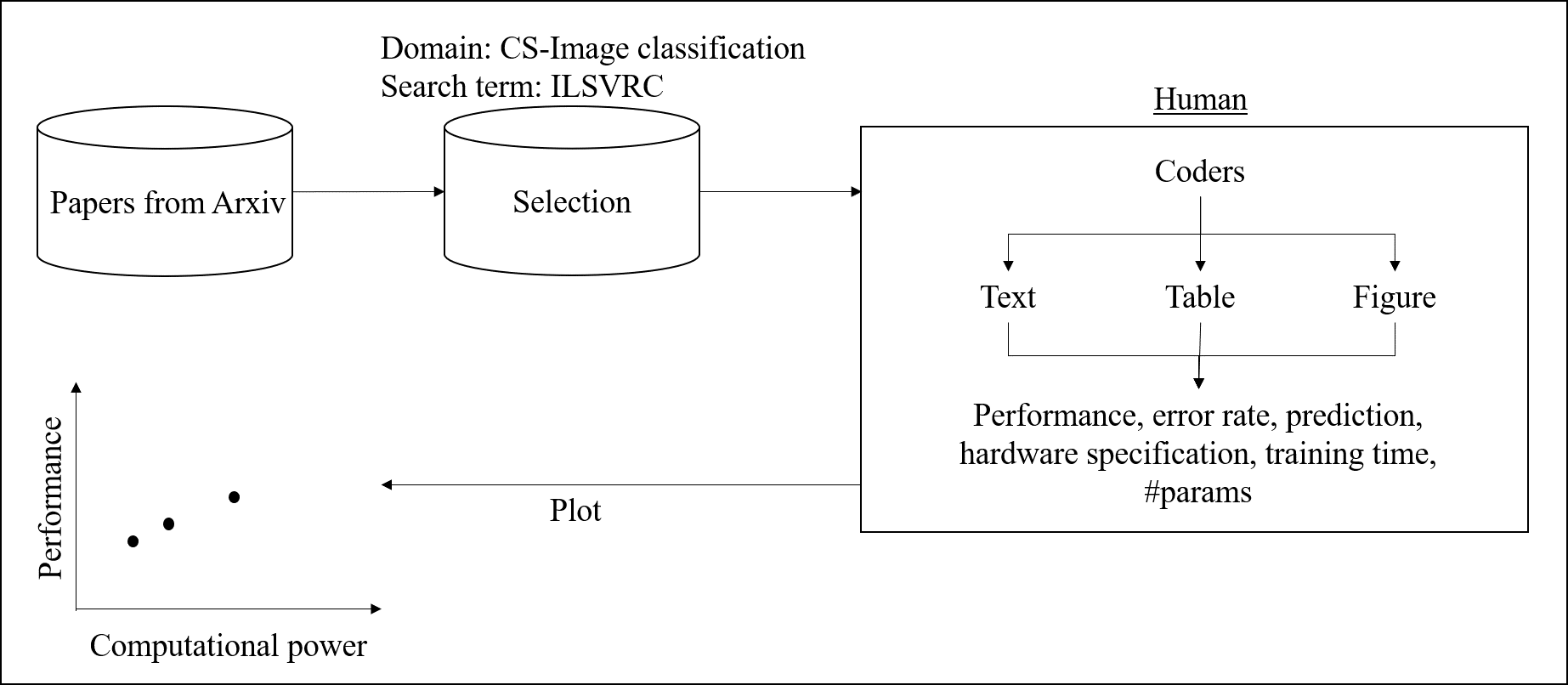}
 \caption{Overview of data collection and extraction process}
 \label{fig:dl_entity_extraction}
\end{figure}

We extract information from these pre-prints using a manual review. Our results were also cross-checked with the data provided by the benchmark website \url{paperswithcode.com}, where authors upload their papers and their respective codes along with the achieved performance metrics of their model (see Figure \ref{fig:dl_entity_extraction} for details) .

Since pre-prints are named using a combination of information such as the paper publication year and month, submission number, and version update, we can automatically extract them when the papers are made public. For example, the Gpipe paper \cite{huang2019gpipe} can be identified by the "\url{1811.06965v5}" identifier indicating that it was made public on November, 2018.

Despite manual review, for many papers we are unable to estimate training compute in consequence of the lack of information provided by their authors.  For example, when the hardware usage data is provided, but model details such as training time is not, it is not possible to estimate the model computing power. Hardware performance data are mostly gathered from external sources such as official hardware designers plataforms (e.g. NVIDIA, Google) or publicly-available databases (e.g. Wikipedia).

\subsection{Application Area: Images}

We examine five applications of deep learning to images: image classification, object detection, face detection, image generation and pose estimation.

\subsubsection{Image classification}
\label{subs:img_classification}
Image classification is a computer vision task where the content of an image is identified using only the image itself. For example, an image classification algorithm performs a set of instructions to calculates the probability that an image contains a cat or not.  There a number of image classification datasets, including Caltech 101 \cite{fei2004caltech}, Caltech 256 \cite{griffin2007caltech}, ImageNet \cite{russakovsky2015imagenet}, CIFAR-10/100 \cite{krizhevsky2014cifar}, MNIST \cite{lecun1998mnist}, SVHN \cite{netzer2019street}, STL-10 \cite{coates2011analysis}, Fashion-MNIST \cite{xiao2017fashion}, CINIC-10 \cite{darlow2018cinic}, Flowers-102, iNaturalist \cite{van2018inaturalist}, EMNIST-Letters \cite{cohen2017emnist}, Kuzushiji-MNIST \cite{lamb2018deep}, Stanford Cars \cite{krause2013collecting}, and MultiMNIST \cite{eslami2016attend}. We focus on ImageNet and CIFAR-10/100 because their long history allows us to gather many data points (i.e. papers).

One simple performance measure for image classification is the share of total predictions for the test set that are correct, called the ``average accuracy.''  (Or, equivalently, the share that are incorrect).  A common instantiation of this is the ``top-k'' error rate, which asks whether the correct label is missing from the top k predictions of the model. Thus, the top-1 error rate is the fraction of test images for which the correct label is not the top prediction of the model. Similarly, the top-5 error rate is the fraction of test images for which the correct label is not among the five predictions.

\noindent\textbf{Benchmark: ImageNet}

ImageNet refers to ImageNet Large Scale Visual Recognition Challenge (ILSVRC). It is a successor of PASCAL Visual Object Classes Challenge (VOC) \cite{everingham2011pascal}. ILSVRC provides a dataset publicly and runs an annual competition as PASCAL VOC. Whereas PASCAL VOC supplied about 20,000 images labelled as $1$ of $20$ classes by a small group of annotators, ILSVRC provides about $1.5M$ images labelled as $1$ of $1000$ classes by a large group of annotators.
The ILSVRC2010 dataset contains $1,261,406$ training images. The minimum number of training images for a class is $668$ and the maximum number is $3047$. The dataset also contains $50$ validation images and $150$ test images for each class. The images are collected from Flickr and other search engines. Manual labelling is crowdsourced using Amazon Mechanical Turk. 

In the ILSVRC2010 competition, instead of deep learning, the winner and the outstanding team used support vector machines (SVM) with different representation schemes. The NEC-UIUC team won the competition by using a novel algorithm that combines SIFT \cite{lowe2004distinctive}, LBP \cite{ahonen2006face}, two non-linear coding representations \cite{zhou2010image, wang2010locality}, and stochastic SVM \cite{lin2011large}. The winning top-5 error rate was $28.2\%$. The second best performance was done by Xerox Research Centre Europe (XRCE). XRCE combined an improved Fisher vector representation \cite{perronnin2007fisher}, PCA dimensionality reduction and data compression, and a linear SVM \cite{perronnin2010improving}, which resulted in top-5 error rate of $33.6\%$. The trend of developing advanced Fisher vector-based methods continued until 2014.

Deep learning systems begin winning ILSVRC in 2012, starting with the SuperVision team from University of Toronto which won with AlexNet, achieving a top-5 error rate of $16.4\%$ \cite{krizhevsky2012imagenet}. Since this victory, the majority of teams submitting to ILSVRCeach year have used deep learning algorithms. 

\noindent\textbf{Benchmark: CIFAR-$10$/$100$}
CIFAR refers to Canadian Institute For Advanced Research (\url{https://www.cifar.ca/}). According to \cite{krizhevsky2009learning}, groups at MIT and NYU collected $80$ million images from the web for building a dataset for unsupervised training of deep generative models. There are two versions of CIFAR dataset, CIFAR-$10$ and CIFAR-$100$, which are subsets of the $80$ million images (\url{https://www.cs.toronto.edu/~kriz/cifar.html}). CIFAR-$10$ contains $6,000$ low-resolution ($32\times32$) color images each of $10$ classes (airplane, car, bird, cat, deer, dog, frog, horse, ship, truck). CIFAR-$100$ dataset contains $600$ low-resolution ($32\times32$) color images each of $100$ classes from $20$ super-classes (aquatic mammals, fish, flowers, food containers, fruit and vegetables, household electrical devices, household furniture, insects, large carnivores, large man-made outdoor things, large natural outdoor scenes, large omnivores and herbivores, medium-sized mammals, non-insect invertebrates, people, reptiles, small mammals, trees, vehicle $1$, vehicles $2$). All the images were annotated by paid students. In 2010, \cite{krizhevsky2010convolutional} trained a two-layer convolutional Deep Belief Network (DBN) on NVIDIA GTX 280 GPU using CIFAR-$10$ dataset. It took $45$ hours to pre-train and $36$ hours to fine-tune and achieved the accuracy rate of $78.90\%$. 

\subsubsection{Object Detection}
Object detection is the task of localization and classification of multiple objects in a single image. Localization means drawing a bounding box for each object. Classification means identifying the object in each bounding box. Localization becomes instance segmentation if, instead of a bounding box, an object is outlined. Whereas image classification identifies a single object in a single image, object detection identifies multiple objects in a single image using localization.

There are various performance measures for object detection, all based around the same concept. Intersection Over Union (IOU) measures the overlap between two bounding boxes: the ground truth bounding box and the predicted bounding box. This is calculated with a Jaccard Index, which calculates the similarity between two different sets, $A$ and $B$, as $J(A,B) = \frac{|A \cap B|}{|A\cup B|}$.  Thus, IOU is the area of the intersection between the two bounding boxes divided by the area of the union of both two bounding boxes. It is $1$ if the ground truth bounding box coincides with the predicted bounding box.
 
Box Average Precision (AP), which is also called mean average precision (mAP), sums IOUs between $0.5$ and $0.95$ and divides the sum by the number of the IOU values.

\noindent\textbf{Benchmark: MS COCO}
COCO refers to Microsoft Common Objects in COntext (MS COCO) released in $2014$ \cite{lin2014microsoft}. The COCO dataset contains $91$ common object categories with $82$ of them having more than $5000$ labeled instances. In total the dataset has $2.5$M manually labeled objects in $328,000$ images. The objects are grouped into $11$ super-categories and then classified into $91$ common object categories by crowdsourced workers on Amazon’s Mechanical Turk platform.  Like other benchmarks, the COCO dataset is publicly available so that new algorithms can be run on it (\url{http://cocodataset.org/}). \cite{he2016deep} applied Faster R-CNN to COCO dataset on a $8$-GPU computer for $80$k iterations and achieved AP of $34.9$.

\subsubsection{Face Detection}
is the task that determines the location and size of a human face in digital images. Given an image, the goal of facial recognition is to determine whether there are any faces and return the bounding box of each detected face. Other objects like trees, buildings, and bodies are ignored from the digital image. Face detection can be regarded as a specific case of object-class detection, where the task is finding the location and sizes of all objects in an image that belongs to a given class \cite{boesch2021face}. Popular benchmarks are WIDER Face (Hard, Medium, and Easy), FDDB, Annotated Faces in the Wild, and PASCAL Face. 

Face detection is measured using the metric Average Precision (AP) which is a popular metric for evaluating the accuracy of object detectors by estimating the area under the curve (AUC) of the precision $\times$ recall relationship \cite{padilla2020survey}.

\noindent\textbf{Benchmark: WIDER Face (Hard)}
WIDER Face refers to a face detection benchmark dataset, of which images are selected from the publicly available WIDER dataset. According to \cite{yang2016wider}, it has 32,203 images and label 393,703 faces with a high degree of variability in scale, pose and occlusion as depicted in the sample images. WIDER FACE dataset is organized based on 61 event classes. For each event class, it is randomly selected 40\%/1\%/50\% of it data as training, validation and testing sets. It uses the same evaluation metric employed in the PASCAL VOC dataset \cite{Everingham10}. It has three levels of difficulty: ’Easy’, ’Medium’, ’Hard’ based on the detection rate of EdgeBox \cite{zitnick2014edge}. The average recall rates for these three levels are 92\%, 76\%, and 34\%, respectively, with 8K proposal per image.

\subsubsection{Image Generation (Synthesis)} is the process of artificially generating images that contain some particular desired content. It is analogous to the inverse of the classification problem: generating an image that contains the visual contents that are associated with a specific label \cite{KUMAR2020167}. Popular benchmarks are CIFAR-10, ImageNet ($32\times32$ \& $64\times64$), STL-10, and LSUN Bedroom.

Image generation is measured using the Frechet Inception Distance score, or FID for short, which is a metric that calculates the distance between feature vectors calculated for real and generated images. The FID score is used to evaluate the quality of images generated by generative adversarial networks, and lower scores have been shown to correlate well with higher quality images \cite{brownlee_2019}.

\noindent\textbf{Benchmark: CIFAR-$10$}
As already mentioned in \ref{subs:img_classification}, CIFAR refers to Canadian Institute For Advanced Research (\url{https://www.cifar.ca/}). According to \cite{krizhevsky2009learning}, it created for building a dataset for unsupervised training of deep generative models. CIFAR-$10$ contains 6K low-resolution ($32\times32$) color images each of $10$ classes (airplane, car, bird, cat, deer, dog, frog, horse, ship, truck).

\subsubsection{Pose Estimation}
is a computer vision task that responsible for detecting and classifying the joints in the human body. Essentially it is a way to capture a set of coordinates for each joint (arm, head, torso, etc.,) which is known as a key point that can describe a pose of a person. The connection between these points is known as a pair. The connection formed between the points has to be significant, which means not all points can form a pair. From the outset, the aim of pose estimation is to form a skeleton-like representation of a human body and then process it further for task-specific applications \cite{barla}.

Today, the most powerful image processing models are based on convolutional neural networks (CNNs). Hence, state-of-the-art methods are typically based on designing the CNN architecture tailored particularly for human pose inference \cite{odemakinde_2021}.

The most popular benchmarks in this task are MPII Human Pose, MS COCO, Leeds Sports Poses, OCHuman and ITOP top-view.

Pose estimation is measured using the  PCKh-0.5 which is a modified version of PCK (Percentage of Correct Key-points). PCKh is also defined as the head-normalized probability of the correct keypoint metric. In PCKh, joint detection is considered correct if the predicted joint location is with a certain threshold from the true joint location. But the threshold should be adaptively selected based on the individual’s size \cite{andriluka2017pose}.

\noindent\textbf{Benchmark: MPII Human Pose} Refers to a benchmark for evaluation of articulated human pose estimation. The dataset includes around 25K images containing over 40K people with annotated body joints. The images were systematically collected using an established taxonomy of every day human activities. Overall the dataset covers 410 human activities and each image is provided with an activity label. Each image was extracted from a YouTube video and provided with preceding and following un-annotated frames. In addition, for the test set we obtained richer annotations including body part occlusions and 3D torso and head orientations \cite{andriluka14cvpr}.

\subsection{Application area: Text}

Deep Learning has been applied to various text-related tasks, including: named entity recognition, machine translation, question answering, text classification, text generation, text summarization, sentiment analysis, emotion recognition, part-of-speech tagging. In this section, we consider three: entity recognition, machine translation, and question answering.

\subsubsection{Named Entity Recognition}
Named entity recognition is the task of identifying and tagging entities in text with pre-defined classes (also called ``types'').  For example, Amazon Comprehend Medical extract relevant medical information such as medical condition, medication, dosage, strength, and frequency from unstructured text such as doctors’ notes \cite{bhatia2019comprehend}. Popular benchmarks are CoNLL 2003, Ontonotes v5, ACE 2004/2005, GENIA, BC5CDR, SciERC. We focus on CoNLL2003.

Named Entity Recognition is measured using an F1 score, which is the harmonic mean of the precision and the recall on that task. The precision is the percentage of named entities discovered by an algorithm that are correct. The recall is the percentage of named entities present in the corpus that are discovered by the algorithm. Only an exact match is counted in both precision and recall.  The F1 score goes to 1 only if the named entity recognition has perfect precision and recall, that is, it finds all instances of the classes and nothing else.

\noindent\textbf{Benchmark: CoNLL2003}
\cite{sang2003introduction} shared the CoNLL2003 dataset for language-independent named entity recognition of the following classes: people, locations, organizations and names of miscellaneous entities. The dataset consists of a training file, a development file, a test file, and a large file with unlabeled data in each of English and German. The four files in English are taken from the Reuters Corpus (\url{http://www.reuters.com/researchandstandards/}). The English training file has $203{,}621$ tokens from $14{,}987$ sentences across $946$ articles. The English test file has $7{,}140$  location tokens, $3{,}438$ miscellaneous entity tokens, $6{,}321$ organization tokens, and $6{,}600$ types of person tokens. The English development file has $51{,}362$ tokens from $3{,}466$ sentences from $216$ articles. It has $51{,}362$ tokens, including $1{,}837$ locations, $922$ miscellaneous entities, $1{,}341$ organizations, and $1{,}842$ types of people. The English test file has $46{,}435$ tokens from $3{,}684$ sentences in $231$ articles. It has $46{,}435$ tokens, including $1{,}668$ locations, $702$ miscellaneous entities, $1{,}661$ organizations, and $1{,}617$ types of people.

\subsubsection{Machine Translation (MT)}
MT tasks a machine with translating a sentence in one language to that in a different language. MT can be viewed as a form of natural language generation from a textual context. MT can be categorized into rule-based MT, statistical MT, example-based MT, hybrid MT, and neural MT (i.e., MT based on DL). MT is another task that has enjoyed a high degree of improvement due to the introduction of DL. Benchmarks are WMT 2014/2016/2018/2019 \cite{bojar2014findings} and IWSLT 2014/2015 \cite{cettolo2014report}.
 
BLEU (Bilingual Evaluation Understudy) \cite{papineni2002bleu} score is a metric for translation and computes the similarity between human translation and machine translation based on n-gram. An n-gram is a continuous sequence of n items from a given text. The score is based on precision, brevity penalty, and clipping. The modified n-gram precision means the degree of overlap in n-gram between reference sentence and translated sentence. Simply, precision is the number of candidate n-grams which occur in any reference over the total number of n-grams in the candidate. Sentence brevity penalty is a factor that rescales a high-scoring candidate translation by considering the extent to match the reference translations in length, in word choice, and in word order. It is computed by a decaying exponential in the test corpus’ effective reference length ($r$) over the total length of the candidate translation corpus ($c$), \(\frac{r}{c}\). Hence the brevity penalty is $1$ if $c$ exceeds $r$ and \(\exp(1-\frac{r}{c})\) otherwise. 

BLEU is a multiplication of an exponential brevity penalty factor and the geometric mean of the modified n-gram precisions after case folding, as the equation below.
\[
BLEU = \min\{1,e^{(1-r/c)}\} \times e^{(\sum_{n=1}^N w_n \log p_n)}.
\]
Here, $N$ is the maximum number that $n$ can have. $w_n$ is a weight on n-gram. $p_n$ is the modified n-gram precision.

BLEU score ranges from 0 to 1. 1 is the best possible score but is only achieved when a translation is identical to a reference translation. Thus even a human translator may not reach 1. BLEU has two advantages. First, it can be applied to any language. Second, it is easy and quick to compute. BLEU is known to have high correlation with human judgments by computing the average of individual sentence judgment errors over a test corpus.

\noindent\textbf{Benchmark: WMT2014}
WMT`14 contains 4.5M sentence pairs (116M English words and 110M German words) as training data (\url{https://www.statmt.org/wmt14/translation-task.html}). 

\subsubsection{Question Answering (QA)}
QA is a task of machine to generate a correct answer to a question from an unstructured collection of documents in a certain natural language. QA requires reading comprehension ability. Reading comprehension of a machine is to understand natural language and to comprehend knowledge about the world.  

Benchmarks include but are not limited to Stanford Question Answering Dataset (SQuAD), WikiQA, CNN, Quora Question Pairs, Narrative QA, TrecQA, Children’s Book Test, TriviaQA, NewsQA, YahooCQA.

F1 score and Exact Match (EM) are popular performance measures. EM measures the percentage of predictions that match any one of the ground truth answers exactly. The human performance is known to be 82.304 for EM and 91.221 for F1 score.

\noindent\textbf{Benchmark: SQuAD1.1}
SQuAD consists of questions posted by crowd workers on a set of Wikipedia articles. And the answer to every question may be in a segment of text. SQuAD1.1 contains 107,785 question-answer pairs on 536 articles  \cite{rajpurkar2016squad}. \cite{rajpurkar2016squad} collected question-answer pairs by crowdsourcing using curated passages in top 10,000 articles of English Wikipedia from Project Nayuki’s Wikipedia’s internal PageRanks. Plus, the authors collected additional answers to the questions that have crowdsourced answers already. 

The dataset is available here  (https://worksheets.cdalab.org/worksheets/0xd53d03a48e f64b329c16b9baf0f99b0c/).

\subsection{Application area: Sound}

\subsubsection{Speech recognition}
Speech recognition is the task of recognizing speech within audio and converting it into the corresponding text. The first part of recognizing speech within audio is performed by an acoustic model and the second part of converting recognized speech into the corresponding text is done by a language model \cite{jelinek1976continuous}. The traditional approach is to use Hidden Markov Models (HMMs) and Gaussian Mixture Models (GMMs) for acoustic modeling in speech recognition. Artificial neural networks (ANNs) were applied to speech recognition since the 1980s. ANNs empowered the traditional approach from the end of 20th century \cite{hochreiter1997long}. Recently, DL models such as CNNs and RNNs improved the performance of acoustic models and language models, respectively. More recently, end-to-end automatic speech recognition based on CTC (Connectionist Temporal Classification) \cite{graves2006connectionist,graves2014towards} has been growing in popularity. Baidu's DeepSpeech2 \cite{amodei2016deep} and Google’s LAS (Listen, Attend and Spell) \cite{chan2016listen} are examples. Speech recognition also requires large scale high quality datasets in order to improve performance. 

One simple way to measure the performance of speech recognition is to compare the body of text read by a speaker with the transcription written by a listener. And, WER (Word Error Rate) is a popular metric of the performance of a speech recognition. It is difficult to measure the performance of speech recognition because the recognized word sequence can have a different length from the reference word sequence. WER is based on Levenshtein distance in word level. In addition, dynamic string alignment is utilized to cope with the problem of the difference in word sequence lengths. WER can be computed by dividing the sum of the number of substitutions, the number of deletions, the number of insertions by the number words in a reference sequence. Namely, the corresponding accuracy can be calculated by subtracting WER from 1. The exemplary benchmarks are Switchboard, LibriSpeech, TIMIT, and WSJ.

\noindent\textbf{Benchmark: ASR SWB Hub500}
Automatic Speech Recognition Switchboard Hub 500 (ASR SWB Hub500) dataset contains 240 hours of English telephone conversations collected by Linguistic Data Consortium (LDC) \cite{ld20022000} (\url{https://catalog.ldc.upenn.edu/LDC2002S09}).

\section{Model Analysis}\label{supp:model_analysis}

\subsection{Estimating Hardware Burden from Network Operations}\label{supp:network_ops}

To find all the data needed to estimate the computing power used to train a model can be quite challenging. This is mainly due to factors such as:

\begin{enumerate}
  \item[1] Data Scarcity;
  \item[2] Lack of accurate data;
  \item[3] Data Inconsistency;
\end{enumerate}

In terms of (1), in many papers only portions of the data necessary for us to proceed with the estimates are reported. In relation to (2), some data, such as training time, are not reported very precisely, thus corroborating the increase of residuals in our model. Still about (2), authors generally do not report the hardware computing precision used in their experiments. While there are some ways to make good guesses about what they use, there is some residual uncertainty. Lastly, (3), sometimes authors conflate multiply-add and floating-point operations (though sometimes these are in fact the same when the hardware executes fused multiply-add operations).

\begin{figure}[!htb]
 \centering
 \includegraphics[width=\columnwidth]{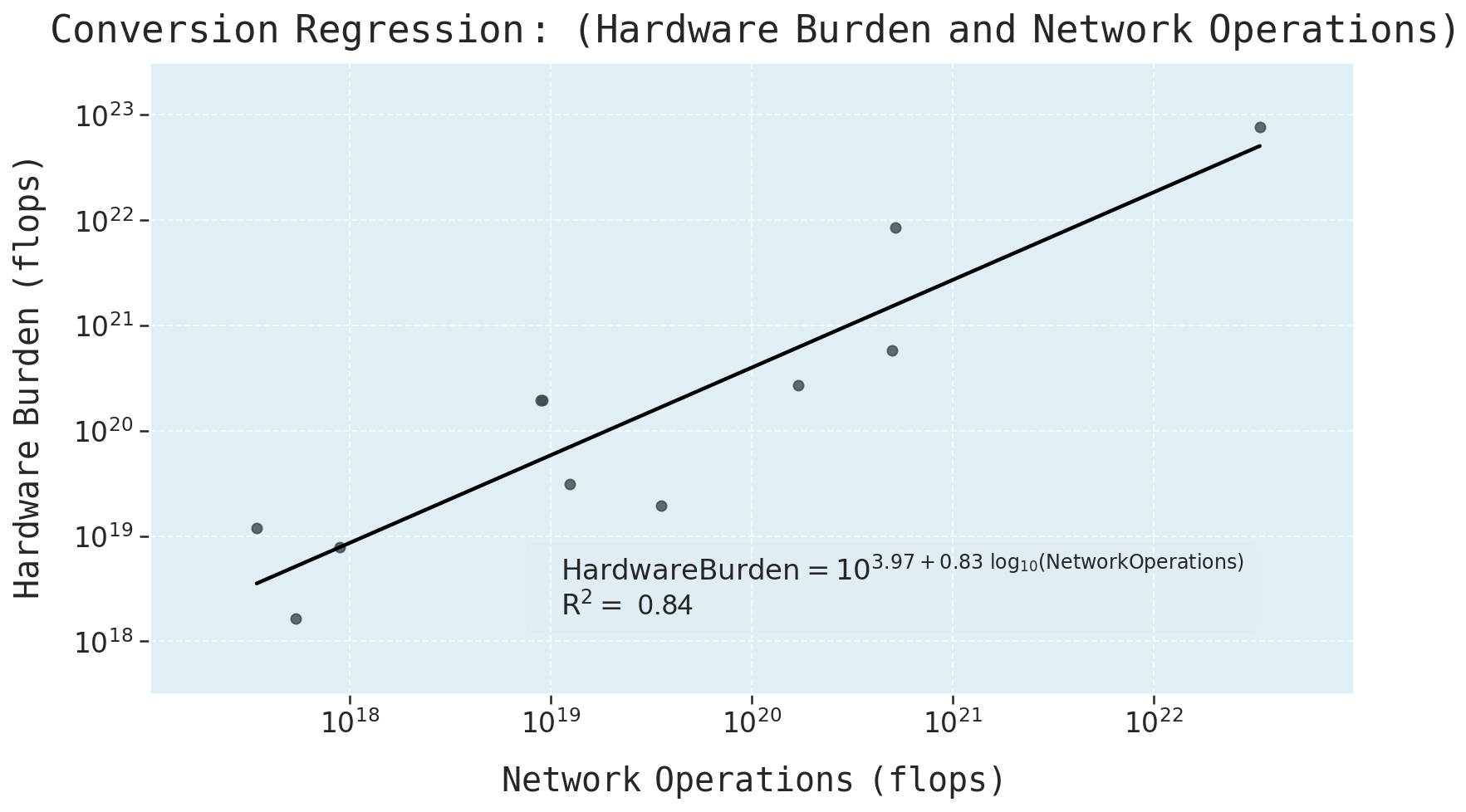}
 \caption{Conversion regression between hardware burden and network operations}
 \label{fig:hw_burden_conversion}
\end{figure}

In our analysis we have Hardware Burden as our main computing power metric. Alternatively, we can also estimate the computing power by looking at the total number of operations performed by the neural network (a.k.a Network Operations). As reported in table \ref{tab:data_gathering}, for some models belonging to the ImageNet benchmark, it was possible to estimate computing power through these two metrics. Consequently, we performed a linear regression model (see Figure \ref{fig:hw_burden_conversion}) so that we could more accurately understand the relationship between these two variables. Given its high statistical significance (p-value $< 0.01$), as well as its high $R^2$ ($0.84$), this model was used to convert Network Operations into Hardware Burden for those models where it was possible to estimate the value for Network Operations only \footnote{Considering that this same approach would be impossible to be performed for the other benchmarks, we used the same ImageNet model to perform the conversion between these two variables in the MS COCO benchmark.}.

\subsection{Quantile regression analysis}\label{supp:quantile}

\begin{figure}[!htb]
 \centering
 \includegraphics[width=\columnwidth]{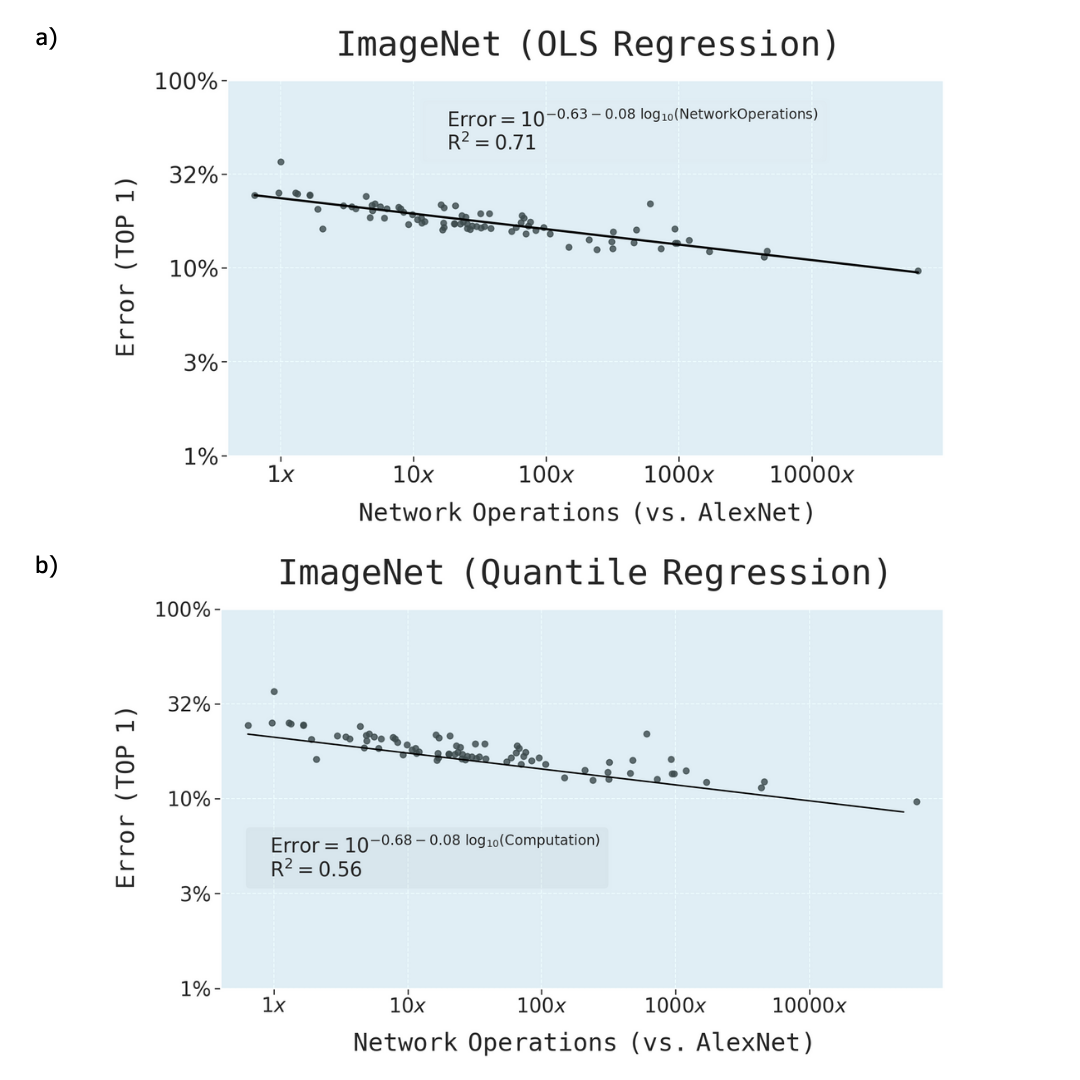}
 \caption{Comparison of conditional mean and 10\% quantile regressions}
 \label{fig:imagenet_quantile_ols}
\end{figure}

Figure \ref{fig:imagenet_quantile_ols} shows a comparison between the conditional-mean regression shown in the paper to a quantile regression ($10\%$) which better approximates the best performance possible for any level of computational burden.  

\section{Regression analog example}\label{supp:reg}

Consider the following generative $d$-dimensional linear model: $y(x) = \theta^T x + z$, where $z$ is Gaussian noise. Given $n$ independent $(x,y)$ samples, the least squares estimate of $\theta$ is $\hat{\theta}_{LS} = (X^T X)^{-1} X^T Y$, yielding a predictor $\hat{y}(x_0) = \hat{\theta}_{LS}^T x_0$ on unseen $x_0$.\footnote{$X \in \mathbb{R}^{n \times d}$ is a matrix concatenating the samples from $x$, and $Y$ is a $n$-dimensional vector concatenating the samples of $y$.} The root mean squared error of this predictor can be shown\footnote{When both $x$ and $x_0$ are drawn from an isotropic multivariate Gaussian distribution.} to scale as $O\left(\sqrt{\frac{d}{n}}\right)$. Suppose that $d$ (the number of covariates in $x$) is very large, but we expect that only a few of these covariates (whose identities we don't know) are sufficient to achieve good prediction performance. A traditional approach to estimating $\hat\theta$ would be to use a small model, i.e. choosing only some small number of covariates, $s$, in $x$, chosen based on expert guidance about what matters.  When such a model correctly identifies all the relevant covariates (the ``oracle" model), a traditional least-squares estimate of the $s$ covariates is the most efficient unbiased estimate of $\theta$.\footnote{Gauss-Markov Theorem.}  When such a model is only partially correct and omits some of the relevant covariates from its model, it will quickly learn the correct parts as $n$ increases but will then have its performance plateau. An alternative is to attempt to learn the full $d$-dimensional model by including all covariates as regressors. Unfortunately, this flexible model is often too data inefficient to be practical. 

Regularization can help. In regression, one of the simplest forms of regularization is the Lasso \cite{tibshirani1996regression}, which penalizes the number of non-zero coefficients in the model, making it sparser. Lasso regularization improves the root mean squared error scaling to $O\left(\sqrt{\frac{s \log d}{n}}\right)$ where $s$ is the number of nonzero coefficients in the true model \cite{meinshausen2009lasso}. 
Hence if $s$ is a constant and $d$ is large, the data requirements of Lasso is within a logarithmic factor of the oracle model, and exponentially better than the flexible least squares approach.
This improvement allows the regularized model to be much more flexible (by using larger $d$), but this comes with the full computational costs associated with estimating a large number ($d$) of parameters. 
Note that while here $d$ is the dimensionality of the data (which can be quite large, e.g. the number of pixels in an image), one can also view deep learning as mapping data to a very large number of nonlinear features. If these features are viewed as $d$, it is perhaps easier to see why one would want to increase $d$ dramatically to achieve flexibility (as it would now correspond to the number of neurons in the network). 

To see these trade-offs quantitatively, consider a generative model that has $10$ non-zero parameters out of a possible $1000$, and consider $4$ models for trying to discover those parameters:
\begin{itemize}
 \item \textbf{Oracle model:} has exactly the correct $10$ parameters in the model
 \item \textbf{Expert model:} has exactly $9$ correct and $1$ incorrect parameters in the model
 \item \textbf{Flexible model:} has all $1000$ potential parameters in the model and uses the least-squares estimate
 \item \textbf{Regularized model:} like the flexible model, it has all $1000$ potential parameters but now in a regularized (Lasso) model
\end{itemize}
We measure the performance as $-\log_{10}(RMSE)$, where $RMSE$ is the normalized root mean squared error between the prediction computed using the estimated parameters and the prediction computed using the true 1000-dimensional parameter vector. The prediction MSE is averaged over query vectors sampled from an isotropic Gaussian distribution.

\begin{figure}[!htb]
 \centering
 \includegraphics[width=\columnwidth]{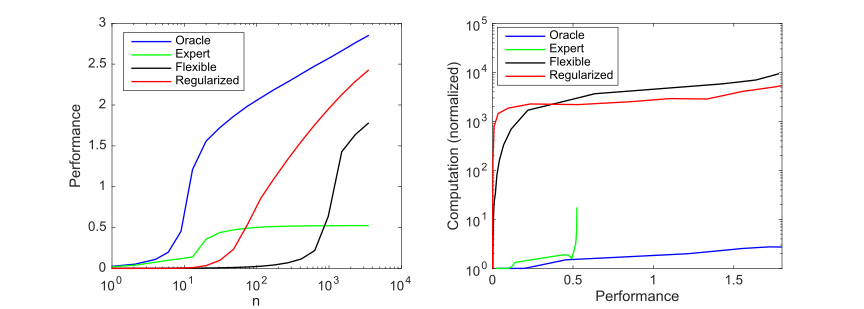}
 \caption{The effects of model complexity and regularization on model performance (measured as the negative $\log_{10}$ of normalized root mean squared error of the prediction compared to the optimal predictor)
 and on computational requirements, averaged over $1000$ simulations per case.  \textbf{(a)} Average performance as sample sizes increase.  \textbf{(b)} Average computation required to improve performance.}
 \label{fig:theory_template}
\end{figure}

As Figure \ref{fig:theory_template}(a) shows, the neural-network analog (the flexible, regularized model) is much more efficient with data than an unregularized flexible model, but considerably less so than the oracle model or (initially) the expert model.  Nevertheless, as the amount of data grows, the regularized flexible model outperforms expert models that don't capture all contributing factors. This graph generalizes an insight attributed to Andrew Ng: that traditional machine learning techniques do better when the amount of data is small, but that flexible deep learning models do better with more data \cite{kruup_2018},\footnote{In fact sufficiently large neural networks are \emph{universal function approximators} \cite{hornik1989multilayer} implying maximum flexibility.}.  Indeed this is a more-general phenomenon of flexible models having greater potential, but also having vastly greater data and computational needs.\footnote{Another demonstration of this comes from the fact that certain types of deep neural networks can provably be replaced by Gaussian process models that are also flexible and have the advantage of being less black-box, but scale their computational needs even more poorly that neural networks \cite{novak2018}.} In our illustration in Figure \ref{fig:theory_template}, for example, $1{,}500$ observations are needed for the flexible model to reach the same performance as the oracle reaches with $15$.  Regularization helps with this, dropping the data need to $175$.  But, while regularization helps substantially with the pace at which data can be learned from, it helps much less with the computational costs, as Figure \ref{fig:theory_template}(b) shows.

Hence, by analogy, we can see that deep learning performs well because it uses overparameterization to create a highly flexible model and uses (implicit) regularization to make the sample complexity tractable.  At the same time, however, deep learning requires vastly more computation than more efficient models.  Thus, the great flexibility of deep learning inherently implies a dependence on large amounts of data and computation.

\begin{table}[!htb]
\centering
\caption{Hardware Burden Regressions}
\resizebox{\columnwidth}{!}{
\def\arraystretch{1.8}%
\begin{tabular}{|c|c|c|c|}
\hline
 \textbf{Task}  &  \textbf{Dataset}  &  \textbf{Estimated\ \ Regression} &  \textbf{$R^2$} \\ \hline \hline 
  & & &\\
 Image\ \ Classification &  Imagenet &  $\log_{10}\left({\frac{1}{1\ -\ \left(\frac{TOP\ 1}{100}\right)}}\right) = -1.12 + 0.09\ \log_{10}(HardwareBurden)$ & $0.68$ \\ 
  & & &\\

 \hline 
 & & &\\
 Object\ \ Detection &  MS\ \ COCO &  $\log_{10}\left({\frac{1}{1\ -\ \left(\frac{BOX\ AP}{100}\right)}}\right) =  -0.96 + 0.06\ \log_{10}(HardwareBurden)$ & $0.81$ \\
  & & &\\
 
 \hline 
  & & &\\
 Question\ \ Answering &  SQuAD\ 1.1 &  $\log_{10}\left({\frac{1}{1\ -\ \left(\frac{EM}{100}\right)}}\right) = -1.12 + 0.09\ \log_{10}(HardwareBurden)$ & $0.87$ \\ 
  & & &\\
 \hline 
  & & &\\
 Named\ \ Entity\ \ Recognition &  CoNLL\ 2003 &  $\log_{10}\left({\frac{1}{1\ -\ \left(\frac{F1-score}{100}\right)}}\right) = 0.63 + 0.03\ \log_{10}(HardwareBurden)$ & $0.43$ \\
  & & &\\
\hline

\end{tabular}
}
\label{tab:regressions}
\end{table}

\newpage





\section*{Data and code availability}

This paper uses data mainly from arXiv publications and Github repositories. All code for data cleaning and analysis associated with this current submission will be available at \url{https://github.com/MIT-FutureTech/TheComputationalLimitsOfDeepLearning} (under construction). Any updates will also be published on our website, \url{www.computerprogress.com}.

\end{document}